\documentclass[10pt,journal,compsoc,twocolumn ]{IEEEtran}

\usepackage{amsfonts}
\usepackage{mathrsfs}
\usepackage{graphicx}
\usepackage{amsmath}
\usepackage{amssymb}
\usepackage{multicol}
\usepackage{multirow}
\usepackage{threeparttable}

\usepackage{epstopdf}

\usepackage{times}
\usepackage{epsfig}
\usepackage{bbm}
\usepackage{rotating}
\usepackage{array}
\allowdisplaybreaks[4]

\usepackage{subfigure}
\usepackage{bm}
\usepackage{algorithm}
\usepackage{algorithmic}

\begin{document}

\title{Deep Face Recognition with Clustering based Domain Adaptation }

\author{Mei Wang, Weihong Deng
\thanks{Mei Wang and Weihong Deng are with the Pattern Recognition and Intelligent System Laboratory, School of Artificial Intelligence, Beijing University of Posts and Telecommunications, Beijing, 100876, China. E-mail: \{wangmei1,whdeng\}@bupt.edu.cn. (Corresponding Author: Weihong Deng)}}

\maketitle

\begin{abstract}

Despite great progress in face recognition tasks achieved by deep convolution neural networks (CNNs), these models often face challenges in real world tasks where training images gathered from Internet are different from test images because of different lighting condition, pose and image quality. These factors increase domain discrepancy between training (source domain) and testing (target domain) database and make the learnt models degenerate in application. Meanwhile, due to lack of labeled target data, directly fine-tuning the pre-learnt models becomes intractable and impractical. In this paper, we propose a new clustering-based domain adaptation method designed for face recognition task in which the source and target domain do not share any classes. Our method effectively learns the discriminative target feature by aligning the feature domain globally, and, at the meantime, distinguishing the target clusters locally. Specifically, it first learns a more reliable representation for clustering by minimizing global domain discrepancy to reduce domain gaps, and then applies simplified spectral clustering method to generate pseudo-labels in the domain-invariant feature space, and finally learns discriminative target representation. Comprehensive experiments on widely-used GBU, IJB-A/B/C and RFW databases clearly demonstrate the effectiveness of our newly proposed approach. State-of-the-art performance of GBU data set is achieved by only unsupervised adaptation from the target training data.
\end{abstract}


\section{Introduction}

Benefiting from convolutional neural networks (CNNs) \cite{krizhevsky2012imagenet,simonyan2014very,szegedy2015going,he2016deep,hu2017squeeze},
deep face recognition (FR) has been the most efficient biometric technique for identity authentication and has been widely used in enormous areas such as military, finance, public security as well as our daily life. However, deep networks which perform perfectly on benchmark datasets may fail badly on real world applications. This is because the set of real world images is infinitely large and so it is hard for any dataset, no matter how big, to be representative of the complexity of the real world. One persuasive evidence is presented by P.J. Phillips' study \cite{phillips2017cross} which conducted a cross benchmark assessment of VGG model \cite{parkhi2015deep} for face recognition. The VGG model, trained on over 2.6 million face images of celebrities from the Web, is a typical FR systems and achieves 98.95\% on LFW \cite{huang2007labeled} and 97.30\% on YTF \cite{wolf2011face}. However, It only obtains 26\%, 52\% and 85\% on Ugly, Bad and Good partition of GBU database, even if all of images in GBU are nominally frontal.

\begin{figure*}[htbp]
\centering
\includegraphics[width=14cm]{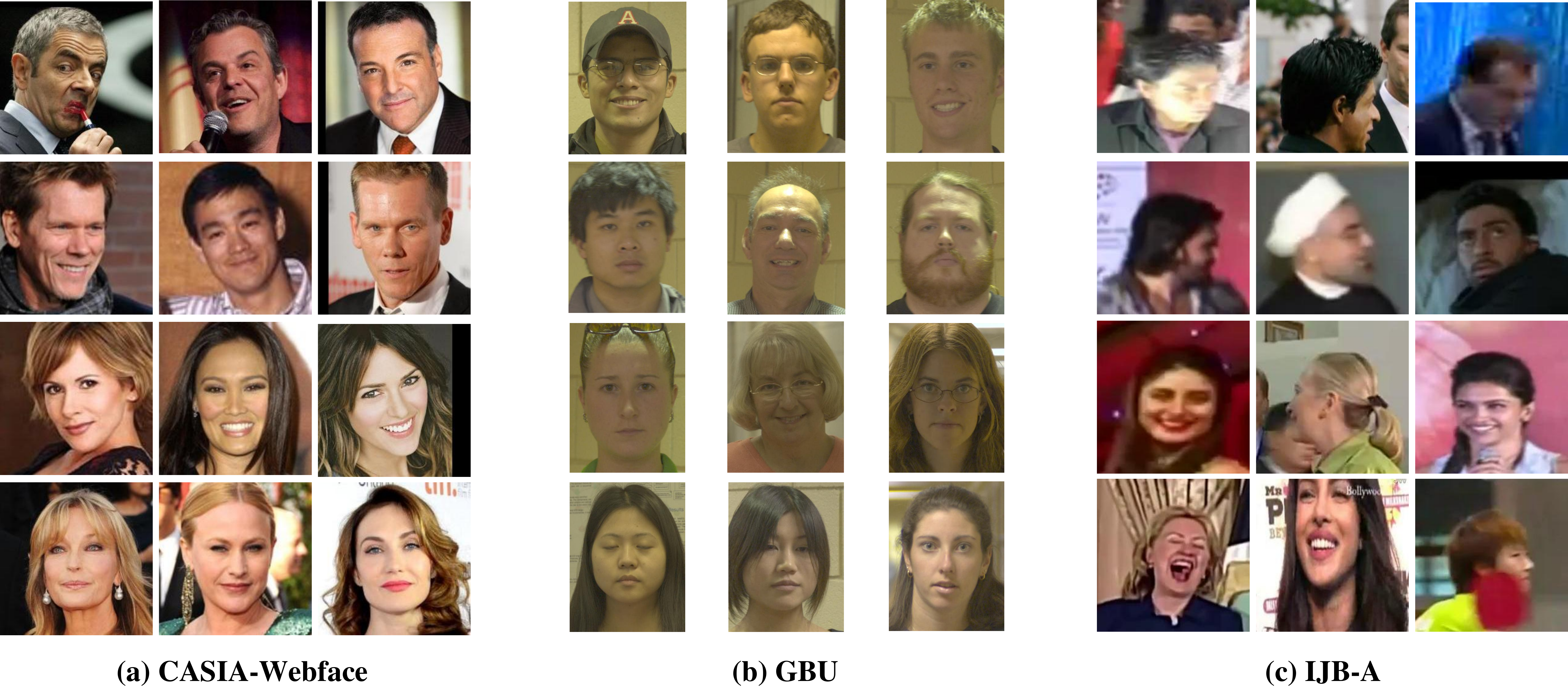}
\caption{ Several sample images of three face databases. From left to right: (a) CASIA-Webface \cite{yi2014learning}, (b) GBU \cite{Phillips2012The} and (c) IJB-A \cite{klare2015pushing}. Compared with CASIA-Webface, GBU is taken outdoors or indoors in atriums and hallways with different lighting conditions; IJB-A covers large pose variations and contains many blurry video frames.}
\label{fig1}
\end{figure*}

The main reason is a different distribution between training data (source domain) and testing data (target domain), referred to as domain or covariate shift. Visual examples of this domain shift are shown in Fig. \ref{fig1}. Each dataset in Fig. \ref{fig1} displays a unique ``signature'' and thus one can easily distinguish them only by these signatures, which proves the existence of significant discrepancies. The images in CASIA-WebFace \cite{yi2014learning} are collected from Internet under unconstrained environment and most of the figures are celebrities and public taken in ambient lighting; The GBU \cite{Phillips2012The} contains still frontal facial images and is taken outdoors or indoors in atriums and hallways with digital camera; IJB-A \cite{klare2015pushing} covers large pose variations and contains many blurry video frames. Sometimes, the images of GBU and IJB-A datasets may be closer to the ones in real life which are taken with digital camera under different shooting environments and contain larger variations.

To alleviate the problems caused by domain shift, the most popular approach is to fine-tune a pre-trained deep network's parameter on testing scenario with the supervision of data label. This straightforward strategy turns out to be problematic because it can be expensive or even infeasible to obtain required amount of labeled data in all possible testing scenarios. Moreover, more and more concerns on privacy may make the collection and human-annotation of the application-collected data become illegal in the future. Fortunately, unsupervised domain adaptation (UDA) is a promising technique aiming to address this problem, which learns a good predictive model for the target (testing) domain using labeled examples from the source (training) domain but only unlabeled examples from the target domain. Recently, many deep UDA methods \cite{wang2018deepdomain} try to learn more transferable representations through mapping both domains into a domain-invariant feature space, and then directly apply the classifier learned from only source labels to target domain, which produce boosted accuracy in various object recognition tasks \cite{Tzeng2014Deep,Long2015Learning,Long2016Deep,Long2016Unsupervised,courty2017optimal,courty2017joint,zellinger2017central,Tzeng2017Adversarial,Ganin2015Unsupervised,pei2018multi,zhang2018collaborative,tzeng2015simultaneous,motiian2017few,ghifary2016deep,bousmalis2016domain}.

In non-deep era, UDA was used for face recognition \cite{ni2013subspace,kan2014domain} in which the distributions of the two datasets are matched by learning a common shared space. In deep era, there have been many well-established deep UDA methods \cite{wang2018deepdomain} for object classification and other computer vision applications. However, most of these methods are not applicable for the face recognition task at all. In particular, face recognition poses two unique challenges for deep UDA different from that in object classification. First, popular methods by the global alignment of source and target domain are no longer sufficient to acquire the discriminating power for classification in deep FR. Second, the face identities (classes) of source and target domain are non-overlapping, so that many skills developed in deep UDA which are used to further improve target performance based on sharing classes are inapplicable. In this sense, designing suitable adaptation method is the key to apply deep face recognition technique in ubiquitous scenes, but few research works has been done in this community.

In this paper, we propose a clustering-based domain adaptation (CDA) method for unconstrained face recognition. In order to address non-overlapping identities between domains, we introduce clustering algorithms into target domain to obtain pseudo-labels, by which the pre-learned model is adapted and the enhanced discriminative representations are learned. Specifically, CDA applies a simplified spectral clustering algorithm which requires neither overlapping classes nor the number of target classes. It generates pseudo-labels through a clustering graph where the nodes represent images and edges signify two images have larger cosine-similarity, and each connected component with at least three nodes in graph is saved as a cluster (identity). This scheme for domain adaptation is fundamentally different from the state-of-the-art methods which generate target pseudo-labels by maximum posterior probability of source classifier \cite{saito2017asymmetric,zhang2018collaborative,chen2018progressive,chen2011co}, these methods can not be utilized in FR due to non-overlapping classes of two domains.

To enhance the quality of clustering-based pseudo-labels, the proposed CDA method applies deep domain confusion network (DDC) \cite{Tzeng2014Deep} and deep adaptation networks (DAN) \cite{Long2015Learning} to conduct global domain alignment before clustering, which optimize the learned representations by minimizing a measure of domain discrepancy, i.e. maximum mean discrepancy (MMD). The hidden representations of images of different domain are embedded in a reproducing kernel Hilbert space, and the mean embeddings of distributions cross domains can be explicitly matched. Through utilizing MMD to optimize pre-learned model, DDC and DAN both alleviate the discrepancy between source and target face database and enhance model performance on target test data. Besides, with more transferable and generalized feature extracted from DDC and DAN, the calculated cosine-similarity of any two target images in our clustering algorithm is more accurate leading to higher quality of pseudo-labels. Comprehensive experiments are carried out in the GBU \cite{Phillips2012The}, IJB-A/B/C \cite{klare2015pushing,whitelam2017iarpa,maze2018iarpa} and RFW \cite{wang2019racial} databases, significant performance gains are reached which indicates the competency of the proposed approach.

Our contributions can be summarized into three aspects.

1) We present a comprehensive study of scene adaptation in face recognition task, and empirically validate the necessity to perform deep domain adaptation. Even the deep models trained by large-scale training Web-collected data still fail to generalize well in many realistic scenes, such as those defined by Ugly data of GBU \cite{Phillips2012The} and the low-quality data of IJB-A dataset \cite{klare2015pushing}. This is caused by the mismatched distribution of training and testing data due to different illuminations, image quality, and shooting angles.

2) We propose a new clustering-based domain adaptation method to address a special domain adaptation task for face recognition where the training (source) and test (target) subjects are non-overlapping. CDA effectively learns the discriminative target feature by aligning the feature domain globally, and, at the meantime, distinguishing the target clusters locally. It first jointly applies DDC and DAN to reduce domain gap and learn domain-invariant representations, and thus provides more reliable underlying face representation for clustering. Then, a simplified spectral clustering method is proposed to generate pseudo-labels in the aligned feature space, and target discriminative representations are learned.

3) We perform extensive face recognition experiments by using the Web-collected dataset \cite{yi2014learning} as source domain, and GBU \cite{Phillips2012The}, IJB-A/B/C databases \cite{klare2015pushing,whitelam2017iarpa,maze2018iarpa} as the target domains, and experimental results demonstrate the superiority of the proposed method. In particular, our method outperforms the state-of-the-art counterparts by a large margin on the GBU dataset, although it is only based on the unsupervised adaptation from the target training data. Moreover, we also utilize our method to perform adaptation across races, and our CDA obtains promising performance on different races of RFW dataset \cite{wang2019racial}.

The remainder of this paper is structured as follows. In the next section, we briefly review related work on deep FR and deep UDA. Then, we introduce the details of MMD and pseudo-labels in Section III. In Section IV, we introduce our clustering based domain adaptation algorithm in detail. Additionally, experimental results are shown and analyzed in Section V. Finally, we conclude and discuss future work.

\section{Related work}

\subsection{Deep face recognition}

In 2014, DeepFace \cite{taigman2014deepface} achieved the state-of-the-art accuracy on the famous LFW benchmark \cite{huang2007labeled}, approaching human performance on the unconstrained condition for the first time, by training a 9-layer model on 4 million facial images. Since then, research of FR focus has shifted to deep-learning-based approaches. More powerful loss functions are explored to learn deep discriminative features and are categorized into Euclidean distance based loss, angular/cosine margin based loss as well as softmax loss and its variations \cite{wang2018deep}. Euclidean distance based loss reduces intra-variance and enlarges inter-variance based on Euclidean distance. DeepID series \cite{wst2008deeply,sun2015deepid3,sun2014deep} combined the face identification (softmax) and verification (contrastive loss) supervisory signals to learn a discriminative representation, and joint Bayesian (JB) was applied to obtain a robust embedding space. They trained 50 networks using a private dataset of 202,595 images and 10,117 subjects. FaceNet \cite{schroff2015facenet} used a triplet loss function aiming to separate the positive pair from the negative one by a distance margin and achieves good performance (99.63\%) on LFW. VGG model \cite{parkhi2015deep} is a typical application based on VGGNet architectures \cite{simonyan2014very}. It was trained on a large scale dataset of 2.6M images of 2622 subjects. Wen et al. \cite{wen2016discriminative} proposed a center loss to reduce the intra-class features variations. To separate samples more strictly and avoid misclassifying the difficult samples, angular/cosine margin based loss is proposed to make learned features potentially separable with a larger angular/cosine distance on a hypersphere manifold, such as Sphereface \cite{liu2017sphereface}, L-softmax \cite{Liu2016Large}, Cosface \cite{wang2018cosface}, AMS \cite{wang2018additive} and Arcface \cite{deng2018arcface}. In addition to Euclidean distance based loss and angular/cosine margin based loss, there are also many works taking effort to normalize feature or weight in softmax loss, e.g. L2-softmax \cite{ranjan2017l2} enforced all the features to have the same L2-norm, so that similar attention is given to good quality frontal faces and blurry faces with extreme pose; Ring loss \cite{zheng2018ring} encouraged norm of samples being value $R$ (a learned parameter) rather than explicit enforcing through a hard normalization operation.

Although these CNN based methods have achieved ultimate accuracy in LFW benchmark, they only focus on utilizing a massive amount of labeled facial images to train a CNN with strong generalization ability and testing on common benchmarks with same distribution. When there is domain shift and it is impossible to obtain labeled data in testing scenarios, the CNN pre-trained on the source data may not generalize well to target data. 

\subsection{Deep unsupervised domain adaptation}

Mimicking the human vision system, domain adaptation is a particular case of transfer learning (TL) that utilizes labeled data in one or more relevant source domains to execute new tasks in a target domain \cite{wang2018deepdomain}. Basically, the main challenge in domain adaptation is the domain shift between the source domain and the target domain. To address this issue, in close-set DA where the images of the source and target domain are from the same set of categories, many UDA approaches are proposed and explore domain-invariant feature spaces by minimizing some measures of domain discrepancy such as statistic loss \cite{Tzeng2014Deep,Long2015Learning,Long2016Deep,Yan2017Mind,Long2016Unsupervised,courty2017optimal}, adversarial loss \cite{Tzeng2017Adversarial,Ganin2015Unsupervised,pei2018multi,zhang2018collaborative,tzeng2015simultaneous,motiian2017few}. MMD is a commonly-used statistic loss for UDA. The DDC proposed by Tzeng et al. \cite{Tzeng2014Deep} is optimized for classification loss in the source domain, while domain difference is minimized by one adaptation layer with the MMD metric. Long et al. \cite{Long2015Learning} proposed DAN that matches the shift in marginal distributions across domains by adding multiple adaptation layers and exploring multiple kernels. 
Adversarial loss makes the distribution of both domains similar enough through domain classifier such that the network is fooled and can be directly used in the target domain. The domain-adversarial neural network (DANN) \cite{Ganin2015Unsupervised} integrated a gradient reversal layer (GRL) to train a feature extractor by maximizing the domain classifier loss and simultaneously minimizing the label predictor loss.

Besides, \cite{saito2017asymmetric,zhang2018collaborative,chen2018progressive,chen2011co,xie2018learning} utilize the pseudo-labels to compensate the lack of categorical information and learn discriminative representations in the target domain. In \cite{saito2017asymmetric}, the idea of tri-training \cite{zhou2005tri} was incorporated into domain adaptation. Two different networks assign pseudo-labels to unlabeled samples, another network is trained by these pseudo-labels to obtain target discriminative representations. Zhang et al. \cite{zhang2018collaborative} iteratively selected pseudo-labeled target samples based on the classifier from the previous training epoch and re-trained the model by using the enlarged training set. 

However, the assumption of close-set DA may not hold in real world application, and the source and target domain may not always share label space. Currently, open-set DA \cite{cao2018partial,zhang2018importance,panareda2017open,liu2019separate,saito2018open} is proposed to address this problem. In open-set DA, different domains only share partial classes and further contain their specific classes. Therefore, the key issue of open-set DA is to separate samples into shared and specific classes and align domains in shared label space. Cao et al. introduced a selective adversarial network (SAN) \cite{cao2018partial} to promote positive transfer by matching the data distributions in the shared label space via splitting the domain discriminator into many class-wise domain discriminators. Separate to Adapt (STA) \cite{liu2019separate} adopted a coarse-to-fine weighting mechanism to progressively separate the samples of unknown and known classes, and used instance-level weights to reject samples of unknown classes in adversarial domain adaptation. Zhang et al. \cite{zhang2018importance} proposed a two domain classifier strategy to identify the importance score of source samples. Satio et al. \cite{saito2018open} proposed a new adversarial learning method in which the feature generator can decrease or increase the probability for specific classes in order to align shared classes or reject specific classes. However, in face recognition, there is no shared class between source and target domain, which is a more complex and realistic setting compared to open-set DA. Domain shift in face recognition can not be addressed through simply aligning domains in shared label space.

\subsection{Unsupervised domain adaptation for face recognition}

In shallow face recognition, many UDA methods \cite{yang2018learning,kan2014domain,ni2013subspace,gheisari2015unsupervised,tao2014sparsity} were utilized to match the distributions of training and testing datasets. Yang et al. \cite{yang2018learning} developed a domain-shared group-sparse dictionary learning model to learn domain-shared representations with aligned joint distributions. 
Kan et al. \cite{kan2014domain} directly converted the source domain data to the target domain in the image space with the help of sparse reconstruction coefficients learnt in the common subspace. Zong et al. \cite{zong2018domain} learned a domain regenerator to regenerate the source and target samples by subspace learning and MMD, such that they can abide by the same or similar feature distributions. Ni et al. \cite{ni2013subspace} sampled several intermediate domains between the source and target domains, and represented each intermediate domain using a dictionary, then they applied invariant sparse codes across these domains to provide a shared feature representation which can be utilized for cross domain recognition. In deep learning era, deeper networks and larger unconstrained images are used to improve the performance of face recognition systems. However, deep FR is still affected by domain shift. Due to the unique challenges of deep FR, very few studies have focused on UDA for deep FR. Luo et al. \cite{Luo2018Adaptation} integrated the maximum mean discrepancies (MMD) estimator to CNN to decrease domain discrepancy. Sohn et al. \cite{sohn2017unsupervised} proposed an UDA method for video FR using large-scale unlabeled videos and labeled still images. They synthesized video frames from images by a set of transformations and utilized images, synthesized images, and unlabeled videos for domain adversarial training. A bi-shifting auto-encoder network (BAE) \cite{kan2015bi} is proposed to enforce the shifted source domain and target domain to share similar distribution, in which each sample of one domain can be sparsely reconstructed by several local neighbors from the other domain. Due to lack of labeled target data, these deep methods only align the feature domain globally, but ignoring the demand of discriminative ability on target domain. It is insufficient for deep FR, which is a fine-grained classification problem. We suggest that pseudo-labels are suitable to address this problem. However, pseudo-label based methods for object classification can not be used in FR because they all assume that there are shared classes between source and target domains and generate target pseudo-labels by maximum posterior probability of source classifier. In this paper, we propose a new clustering-based domain adaptation method to address this unique challenge.

\section{Preliminary}

In our case, we are given a set of labeled data from the source domain, and denote them as $\mathcal{D}_{s}=\{{x^{s}_{i}},{y^{s}_{i}}\}^{M}_{i=1}$, where $x^{s}_{i}$ is the $i$-th source sample, $y^{s}_{i}$ is its category label, and $M$ is the number of source images. A set of unlabeled data from the target domain is given as well and is denoted as $\mathcal{D}_{t}=\{{x^{t}_{i}}\}^{N}_{i=1}$, where $x^{t}_{i}$ is the $i$-th target sample and $N$ is the number of target images. The data distributions of two domains are different, $P(X_s,Y_s)\neq P(X_t,Y_t)$.

\begin{figure*}[htbp]
\centering
\includegraphics[width=17cm]{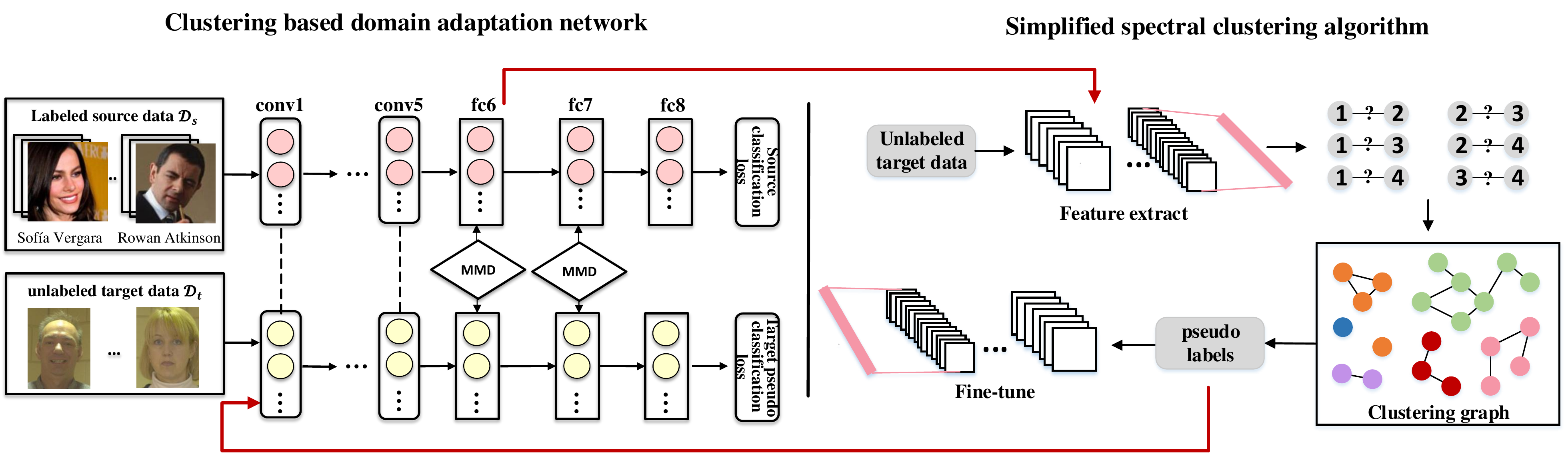}
\caption{ The overall structure of the proposed method. \textbf{Left:} The clustering based domain adaptation network. Source classification loss supervises learning proceeds for source domain; MMD loss aims at minimizing the distribution discrepancy of two domains; target pseudo classification loss aims to learn discriminative target representations on pseudo-labels generated by clustering algorithms. Only using the first two losses to optimize networks is denoted as MMD-based networks. We first train a MMD-based network using labeled source data and unlabeled target data, then utilize target pseudo classification loss to further adapt target CNN after obtaining target pseudo-labels. \textbf{Right:} The simplified spectral clustering algorithm. With the target representations extracted by MMD-based network, a clustering graph is constructed where the nodes represent images and edges signify two images have larger cosine-similarity. Each connected component with at least three nodes is saved as a cluster (identity). Then, we can annotate the clustered nodes with pseudo labels and adapt the target CNN with them. }
\label{fig2}
\end{figure*}

\subsection{Maximum mean discrepancy}

In the field of UDA, MMD \cite{Tzeng2014Deep,Long2015Learning} has been widely adopted as a standard distribution distance metric to measure the discrepancy between source and target domains. Given two distributions $s$ and $t$, the MMD between them is defined as:
\begin{equation}
L_{M}(s,t) = \mathop {\sup }\limits_{{{\left\| \phi  \right\|}_{\cal H}} \le 1} \left\| {{E_{{{\rm{x}}^s} \sim s}}[\phi ({{\rm{x}}^s})] - {E_{{{\rm{x}}^t} \sim t}}[\phi ({{\rm{x}}^t})]} \right\|_{\cal H}^2
\end{equation}
where $E$ represents the expectation with regard to the distribution. $\phi$ represents the function that maps the original data to a reproducing kernel Hilbert space (RKHS). We have $MM{D^2}(s,t)=0$ when $s$ and $t$ share the same distribution based on the statistic tests defined by MMD. The kernel functions which are associated with this mapping, $k(x^{s},x^{t})=\langle\phi(x^{s}),\phi(x^{t})\rangle$, is defined as the convex combination of $m$ PSD kernels ${k_{u}}$,
\begin{equation}
\mathcal{K}=\left \{ k=\sum_{u=1}^{m}\beta _{u}k_{u}:\sum_{u=1}^{m}\beta _{u}=1,\beta _{u}\geq 0,\forall{u}\right \}
\end{equation}
where $\beta _{u}$ is the coefficient of $u$-th kernel and the commonly-used kernel is the Gaussian kernel $k_{u}({x^s},{x^t}) = {e^{ - {{\left\| {{x^s} - {x^t}} \right\|}^2}/\gamma }}$. 
Denote by $\mathcal{D}_{s}=\{{x^{s}_{i}}\}^{M}_{i=1}$ and $\mathcal{D}_{t}=\{{x^{t}_{i}}\}^{N}_{i=1}$ drawn from the distributions $s$ and $t$, respectively, an empirical estimate of MMD is given as:
\begin{equation}
L_{M}(D_s,D_t) = \left\| {\frac{1}{M}\sum\limits_{i = 1}^M {\phi ({\rm{x}}_i^s) - } \frac{1}{N}\sum\limits_{j = 1}^N {\phi ({\rm{x}}_j^t)} } \right\|_H^2 \label{mmd}
\end{equation}

The main idea of MMD-based network, i.e. DDC \cite{Tzeng2014Deep} and DAN \cite{Long2015Learning}, is to integrate MMD estimator to the CNN error so that the domain divergence is minimized. 
However, the formulation of MMD in Eq. (\ref{mmd}) is computed in quadratic time complexity, it is prohibitively time-consuming for deep UDA. Gretton et al. \cite{gretton2012kernel} further suggested an unbiased approximation to MMD with linear complexity and it is suitable for gradient computation in a mini-batch manner:
\begin{equation}
\begin{split}
L_{M}(D_s,D_t)=&\frac{1}{M(M-1)}\sum_{i\neq j}^{M}k(x^{s}_{i},x^{s}_{j})\\
+&\frac{1}{N(N-1)}\sum_{i\neq j}^{N}k(x^{t}_{i},x^{t}_{j})\\
-&\frac{2}{MN}\sum_{i,j=1}^{M,N}k(x^{s}_{i},x^{t}_{j}) \label{mmd2}
\end{split}
\end{equation}

Through optimizing networks by MMD, the final classification decisions are made based on features that are invariant to the change of domains, i.e., have the same or very similar distributions in the source and the target domains, thus, the models trained on source data can generalize to target data.

\subsection{Pseudo label}

Pseudo-label is an alternative method for deep UDA in object classification assuming that source and target domain share the same classes \cite{saito2017asymmetric,zhang2018collaborative,chen2018progressive,chen2011co,xie2018learning}. CNN is trained supervised with source labeled data and is fine-tuned with target pseudo-labeled data that can be obtained by following steps. We denote $\{p_c(x_i^t)|_{c=1}^{m_c}\}$ as the output from the Softmax layer of the source classifier in CNN, where each $p_c(x_i^t)$ is the probability that target sample $x_i^t$ belongs to the $c$-th classes, and $m_c$ is the total number of classes. Then, the pseudo-label of $x_i^t$ can be obtained by choosing the class with the maximum posterior probability:
\begin{equation}
\hat{y_{i}^{t}}=arg \max\limits_{c}\ p_c(x_{i}^{t}) \label{pseudo}
\end{equation}
After that, the network is fine-tuned on pseudo-labeled target data with supervision of Softmax loss.

Furthermore, to suppress the negative influence of falsely-labeled samples, some studies are explored modified strategies which progressively select reliable pseudo-labels from the most confident predictions and re-train the model by using the enlarged training set. It can be formulated as follow:
\begin{equation}
\forall x_i^t \in D_k^t|_{k=1}^{m_c},\ \omega _i=\left\{
             \begin{array}{lll}
             1, & if \ p_k(x_{i}^{t})>\eta  & \\
             0, & otherwise& \\
             \end{array}
\right.
\end{equation}
where $D_k^t|_{k=1}^{m_c}$ denotes the unlabeled target samples $D_t$ are partitioned into $m_c$ classes. $\omega _i=1$ indicates $x_i^t$ to be selected in current training process; otherwise, $x_i^t$ is not to be selected. $\eta$ is a threshold which constrains the maximum posterior probability (confidence) of selected samples. $\eta$ can be a constant, or a variable of the training step \cite{chen2018progressive,xie2018learning}, or a variable of the classification accuracy of the current classifier measured by the labeled source data \cite{zhang2018collaborative}.

\section{Clustering based domain adaptation}

Due to the absence of labeled target samples, most deep DA methods for object classification, such as MMD, only align source and target domain globally. It is not effective enough and cannot ensure accuracy on the target domain in FR tasks where discriminative target representations are required. When lacking of target categorical information, we suggest that pseudo-labels \cite{lee2013pseudo} are suitable to address this problem, which encourages a low-density separation between classes in the target domain. However, adopting UDA in face recognition is a special domain adaptation task where the training (source) and test (target) subjects are non-overlapping, which means that traditional pseudo-labels based UDA methods relying on shared categories are inapplicable. To address this problem, we propose to introduce clustering algorithms into UDA. Many clustering algorithms are feasible for generating pseudo-labels in our clustering-based domain adaptation (CDA) network, and we design a simplified spectral clustering algorithm which is simple but effective for clustering faces in deep feature space. It clusters faces through connected subgraphs and can be adopted even if the number of target classes is large but unknown. The overall architecture of our method is depicted in Fig. \ref{fig2}. 

\subsection{Clustering algorithm}

In this section, we formally introduce the detailed steps of simplified spectral clustering algorithm:

\textbf{Compute similarity matrix.} We feed unlabeled target data $X_{t}$ into a deep model as input and extract deep features $\mathcal{F}(X_{t})$. As we know, the clustering results depend not only on the choice of clustering algorithm, but also on the quality of the underlying face representation. Considering domain shift, the underlying target representation will not be perfect even using a strong source model. Therefore, the deep model here is pre-trained on source samples and further optimized by MMD to improve performance in target domain as much as possible. Then, with these deep presentations, we construct a $N \times N$ similarity matrix, where $N$ is the number of faces in target domain and entry at $(i,j)$, i.e. $s(i,j)$, is the cosine similarity between target representations $\mathcal{F}(x^{t}_{i})$ and $\mathcal{F}(x^{t}_{j})$.

\textbf{Build clustering graphs.} We consider two faces belonging to one identity if their cosine similarity is large. Thus, we can build a clustering graph $\mathcal{G}(n,e)$ according to similarity matrix, where the node $n_{i}$ represents $i$-th target image and edge $e(n_{i},n_{j})$ signifies these two target images have larger cosine-similarity:
\begin{equation}
e(n_{i},n_{j})=\left\{
             \begin{array}{lll}
             1, & if \ s(i,j)>\alpha & \\
             0, & otherwise& \\
             \end{array}
\right. \label{cluster}
\end{equation}
where $\alpha$ is the threshold for edges. Then, we simply save each connected component with at least $p$ nodes as a cluster (identity) and the remaining images will be treated as scattered points. We choose a minimum component size $p=3$. Because the connected components with only one or two nodes may be the ones clustered incorrectly; even if this cluster is correct, low-shot class would deteriorate the long tail distribution of data. Furthermore, the threshold $\alpha$ is vital for clustering. If $\alpha$ is set to be lower, more faces of different identities will be clustered together which contains severe intra-class noise; otherwise, faces of one identity will split into more scattered points and be discarded, or they will split into smaller clusters leading to severe inter-class noise.

\textbf{Pick up scattered points.} Due to large variations, some images can not be clustered and be treated as scattered points. We pick up these scattered points by assuming that all samples of a given identity can be clustered around its corresponding prototype. The prototypes are computed by the average representation $\mu_{k}^t$ of all target samples in one cluster $k$ obtained by connected component:
\begin{equation}
\mu_{k}^t=\frac{1}{\left | D_k^t \right |}\sum_{x_{i}^{t}\in D_k^t}\mathcal{F}(x_{i}^{t})
\end{equation}
where $D_k^t$ is the set of all target images in $k$-th cluster. Then, for each scattered point $x_{i(scatter)}^{t}$, we compute its cosine similarities with all prototypes, and add it to corresponding prototype with the largest cosine similarity. 
To obtain the samples with high confidence, we constrain that the similarity scores should above a certain threshold $\beta$:
\begin{equation}
\begin{split}
&\hat{y_{i}^{t}}=\left\{
             \begin{array}{lll}
             arg \max\limits_{k}\ s_{k}, & if \max\limits_{k}\ s_{k}>\beta  & \\
             \infty , & otherwise& \\
             \end{array}
\right.\\
& {\rm where,} \ s_{k}=cos \left ( \mathcal{F}(x_{i(scatter)}^{t})  ,\mu_{k}^t\right ) \label{scattered}
\end{split}
\end{equation}

So, we only cluster images with higher confidence to alleviate negative influence caused by falsely-labeled samples. Finally, we can annotate all clustered nodes with pseudo label $\hat{y_{i}^{t}}$, and adapt the network with supervision of Softmax loss.

\subsection{Adaptation networks}

We extend the VGGNet \cite{simonyan2014very} and RseNet \cite{he2016deep} architecture to our CDA network. 
As shown in Fig. \ref{fig2}, the architecture of our CDA consists of a source and target CNN, with shared weights. MMD estimators are adopted on higher layers of network which are called adaptation layers. We simply use a fork at the top of the network, after the adaptation layer. The inputs of source CNN are source labeled images while those of target CNN are target unlabeled data. The goal of our approach is to minimize the following loss function:
\begin{equation}
L=L_{S}(X_{s},y_{s})+\lambda \sum_{l\in\mathcal{L} }L_{M}(D_s^l,D_t^l)+L_{T}(X_{t},\hat{y_{t}})
\end{equation}
where the hyperparameter $\lambda$ is a penalty parameter. $\mathcal{D}_*^l$ is the $l$-th layer hidden representation for the source and target examples, and $L_{M}(D_s^l,D_t^l)$ (Eqn. \ref{mmd2}) is the MMD between the source and target evaluated on the $l$-th layer representation. MMD loss makes the distributions of the source and target similar under the hidden representations. Selecting suitable adaptation layers can significantly enhance the transfer efficiency. According to the observation of \cite{yosinski2014transferable}, the transfer ability drops in higher layers with increasing domain discrepancy and transfer learning method would obtain better performance when transferring higher layers of the deep neural network. In CDA, we adopt multi-kernel MMD on the last two layers. $L_{S}(X_{s},y_{s})$ denotes source classification loss on the source data $X_{s}$ and the ground truth labels $y_{s}$, which guarantees the performance of deep network. The third term, i.e. $L_{T}(X_{t},\hat{y_{t}})$, is our target pseudo classification loss on the target data $X_{t}$ and the pseudo-labels $\hat{y_{t}}$, which learns more discriminative representations for target domain:
\begin{equation}
\begin{split}
&L_{S}(X_{s},y_{s})=-\frac{1}{M}\sum_{i=1}^{M}\sum_{c=1}^{m_c} \mathbf{1}_[{c=y_i^s}]logp_c(x_i^s) \\
&L_{T}(X_{t},\hat{y_{t}})=-\frac{1}{\hat{N}}\sum_{i=1}^{\hat{N}}\sum_{c=1}^{\hat{n}_c} \mathbf{1}_[{c=\hat{y}_i^{t}}]logp_c(x_i^t) \label{DA}
\end{split}
\end{equation}
Here, we utilize Softmax loss (Arcface loss \cite{deng2018arcface}) as our classification loss for source and target domain. In source classification loss $L_{S}(X_{s},y_{s})$, $\mathbf{1}_[{c=y_i^s}]$ is 1 when $c=y_i^s$, otherwise, it is 0; $p_c(x_i^t)$ is the probability that source sample $x_i^s$ belongs to the $c$-th classes, and $m_c$ is the total number of source classes. The definition of target pseudo classification loss $L_{T}(X_{t},\hat{y_{t}})$ is similar to that of source classification loss where $\hat{n}_c$ is the total number of target clusters and $\hat{N}$ is the number of target samples clustered successfully. 

\subsection{Clustering based domain adaptation algorithm}

The entire procedure of our method is depicted in Algorithm \ref{al1}. In the first stage, the baseline model is trained with our source data, i.e. CASIA-Webface \cite{yi2014learning}, so that we can use it as our source CNN. In the second stage, source classification loss and MMD loss are used to optimize MMD-based networks (i.e. DDC \cite{Tzeng2014Deep} and DAN \cite{Long2015Learning}). MMD-based network is conducted by source CNN and weight-shared target CNN and is trained with the unlabeled target data and labeled source data so that the deep features are invariant to the change of domains, i.e., have the same or very similar distributions in the source and the target domains, and the performance of target domain is preliminarily improved. In the third stage, we extract deep features of target samples by MMD-based network, then adopt our clustering algorithms to generate pseudo-labels. Benefiting from better performance of MMD-based networks on target domain, the calculated cosine-similarity of any two target images in our clustering algorithm is more accurate leading to higher quality of pseudo-labels. In the forth stage, we adapt the target CNN on these pseudo-labeled target data with supervision of target pseudo classification loss.
MMD-based networks address huge domain discrepancy to learn transferable representations for FR tasks and provide more reliable underlying face representation for clustering; while pseudo-labels encourage a low-density separation between target classes to learn more discriminative representations for FR tasks.

\begin{algorithm}[htbp]
\caption{ Clustering based domain adaptation algorithms.}
\label{al1}
\footnotesize
\begin{algorithmic}[1]
\REQUIRE ~~\\
Source domain labeled samples $\{{x^{s}_{i}},{y^{s}_{i}}\}^{M}_{i=1}$, and target domain unlabeled samples $\{{x^{t}_{i}}\}^{N}_{i=1}$. Network learning rate $\mu$, hyper parameter $\lambda$, $\alpha$, $\beta$ and $p$, network layer parameters $\Theta$.
\ENSURE ~~\\
Network layer parameters $\Theta$.
\STATE \textbf{\emph{Stage-1:}} //  Pre-train
\STATE Train the baseline model on source labeled data;
\STATE \textbf{\emph{Stage-2:}}  // MMD-adaptation \\
Adapt the network with MMD loss and source classification loss to learn domain-invariant representations and provide more reliable underlying face representation for clustering
\STATE \textbf{Repeat:}
\STATE $j=j+1$
\STATE Update the backpropagation error for $x_i$:\\
$\frac{\partial L^{j}}{\partial {x^{s}_{i}}^{(j)}}=\frac{\partial L_{S}^{j}}{\partial {x^{s}_{i}}^{(j)}}+\lambda \frac{\partial L_{M}^{j}}{\partial {x^{s}_{i}}^{(j)}}$\\
$\frac{\partial L^{j}}{\partial {x^{t}_{i}}^{(j)}}=\lambda \frac{\partial L_{M}^{j}}{\partial {x^{t}_{i}}^{(j)}}$
\STATE Update the network layer parameters $\Theta$:\\
$\Theta ^{j+1}=\Theta ^{j}-\mu ^{j}\frac{\partial L^{j}}{\partial \Theta ^{j}}$\\
$=\Theta ^{j}-\mu ^{j}\left ( \sum_{i=1}^{M}\frac{\partial L^{j}}{\partial {x^{s}_{i}}^{(j)}}\frac{\partial {x^{s}_{i}}^{(j)}}{\partial \Theta ^{j}} +\sum_{i=1}^{N}\frac{\partial L^{j}}{\partial {x^{t}_{i}}^{(j)}}\frac{\partial {x^{t}_{i}}^{(j)}}{\partial \Theta ^{j}} \right )$
\STATE \textbf{Until convergence}
\STATE \textbf{\emph{Stage-3:}} // generate target pseudo labels by clustering algorithm
\STATE Extract deep features of target unlabeled data and compute similarity matrix;
\STATE Build clustering graphs according to Eqn. (\ref{cluster}) and save each connected component with at least $p$ nodes as a cluster;
\STATE Add scattered points to corresponding clusters according to Eqn. (\ref{scattered});
\STATE Annotate all clustered nodes with pseudo label $\hat{y_{i}^{t}}$.
\STATE \textbf{\emph{Stage-4:}}  // Pseudo-adaptation \\
Adapt the network with target pseudo-labels using target pseudo classification loss to learn more discriminative target representations
\STATE \textbf{Repeat:}
\STATE $j=j+1$
\STATE Update the network layer parameters $\Theta$:\\
$\Theta ^{j+1}=\Theta ^{j}-\mu ^{j}\frac{\partial L_{T}^{j}}{\partial \Theta ^{j}}=\Theta ^{j}-\mu ^{j}\left (\sum_{i=1}^{N}\frac{\partial L_{T}^{j}}{\partial {x^{t}_{i}}^{(j)}}\frac{\partial {x^{t}_{i}}^{(j)}}{\partial \Theta ^{j}} \right )$
\STATE \textbf{Until convergence}
\end{algorithmic}
\end{algorithm}

\section{Experiments}

In this section, we evaluate our CDA method on five face recognition benchmarks, i.e. GBU \cite{Phillips2012The}, IJB-A/B/C \cite{klare2015pushing,whitelam2017iarpa,maze2018iarpa} and RFW \cite{wang2019racial}. We will begin with introducing the detailed information and evaluation protocol of the datasets we utilized, followed by illustrating the training details of our experiments and presenting results and analyses.

\subsection{Datasets and Evaluation Protocols}

\textbf{CASIA-WebFace:} CASIA-WebFace dataset \cite{yi2014learning} is a large scale face dataset gathered from Internet. It contains 10,575 subjects and 494,414 images. The large scale of labeled facial data does great help to train CNNs. In our experiments, we adopt this dataset as the source domain data for training the classification network.

\begin{figure}[htbp]
\centering
\includegraphics[width=6cm]{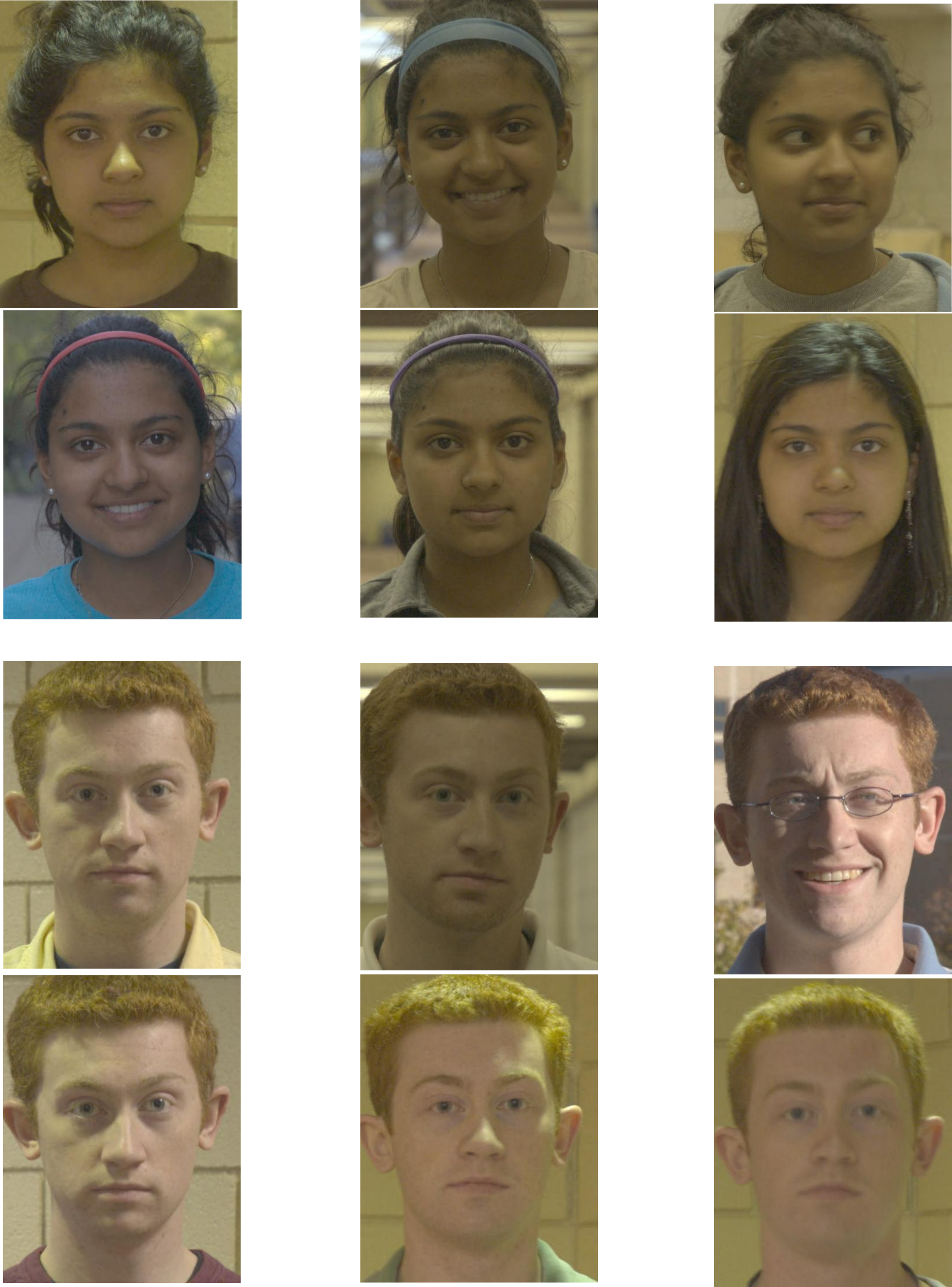}
\caption{ Two example identities of the Good, Bad, and Ugly partition of GBU database. The top two rows show three pairs of images of the same person, sampled from the Good (left), Bad (middle), and Ugly (right) performance conditions. The second two rows show the same type of sample for a second person.}
\label{fig3}
\end{figure}

\textbf{GBU:} Its full name is \emph{The Good, the Bad, and the Ugly Face Challenge} \cite{Phillips2012The}. This dataset consists of three partitions, and different partitions contain pairs of images with different difficulty levels based on the performance of three top performers in the FRVT 2006. The Good partition consists of images which are easy to match; the Bad one contains pairs of average difficulty to recognize; the Ugly one contains pairs considered difficult. Fig. \ref{fig3} shows three pairs of images of each person, sampled from the Good (left), Bad (middle), and Ugly (right) partition. This figure illustrates the variations in the appearance of a person across frontal images, e.g. different settings, expression and hairstyle. Each partition consists of a target set and a query set, and both them contain 1085 images of 437 distinct people. Following the evaluation protocol of \cite{Phillips2012The}, we use receiver operating characteristics (ROC) curve and the verification rate (VR) at a false positive rates (FAR) of 0.001 for each partition to compare the performances of different algorithms. In order to ensure that the subjects in target training set do not appear in target testing set, we utilize part of images from FRGC \cite{phillips2005overview} (without label information) as the target training data, which consists of 19270 still front faces.

\textbf{IJB-A:} IJB-A database \cite{klare2015pushing} contains 5,397 images and 2,042 videos of 500 subjects, which are split into 20,412 frames, 11.4 images and 4.2 videos per subject. It is a joint face detection and FR
dataset, in which both face detection and facial feature point detection are accomplished manually. The key characteristics of IJB-A are that it contains a mixture of images and videos in the wild and covers a full range of pose variations. IJB-A provides 10-split evaluations with two standard protocols, namely, face verification (1:1 comparison) and face identification (1:N search). The performance of verification is reported using the true accept rates (TAR) vs. false positive rates (FAR) (i.e. ROC curve). The performance of identification is reported using the Rank-N (i.e. the cumulative match characteristic (CMC) curve) and the true positive identification rate (TPIR) vs. false positive identification rate (FPIR). There are ten random training (333 subjects) and testing (167 subjects) splits which occur at subject level, using all 500 IJB-A subjects. For each split, we adopt our CDA method by using its training data (without label information) as our target training data and using its testing data as our target testing data. The results are averaged over 10 testing splits.

\textbf{IJB-B:} The IJB-B dataset \cite{whitelam2017iarpa} is an extension of IJB-A \cite{klare2015pushing}, having 1,845 subjects with 21.8K still images (including 11,754 face and 10,044 non-face) and 55K frames from 7, 011 videos.  The dataset is more challenging and diverse than IJB-A, with protocols designed to test detection, identification, verification and clustering of faces. Unlike the IJB-A dataset, it does not contain any training splits. We use images of IJB-A (without label information) as our target training data and use images of IJB-B as our target testing data.

\textbf{IJB-C:} The IJB-C dataset \cite{maze2018iarpa} is a further extension of IJB-B, having 3,531 subjects with 31.3K still images and 117.5K frames from 11,779 videos. In total, there are 23,124 templates with 19,557 genuine matches and 15,639K impostor matches. Similar to IJB-B dataset, the protocols are designed to test detection, identification, verification and clustering of faces. The dataset also contains end-to-end protocols to evaluate the algorithm¡¯s ability to perform end-to-end face recognition. We use images of IJB-A (without label information) as our target training data and use images of IJB-C as our target testing data.

\textbf{RFW:} Racial Faces in-the-Wild (RFW) dataset \cite{wang2019racial} is a testing database for studying racial bias in face recognition. Four testing subsets, namely Caucasian, Asian, Indian and African, are constructed, and each contains about 3000 individuals with 6000 image pairs for face verification. They can be used to fairly evaluate and compare the recognition ability of the algorithm on different races. We use RFW dataset to validate the effectiveness of our CDA method on transferring knowledge across races. In order to perform adaptation experiment, we utilize BUPT-Transferface dataset \cite{wang2019racial} to train our CDA model and test it on RFW. BUPT-Transferface dataset is a training dataset with four race subsets and is released with RFW. One training subset consists of about 500K labeled images of 10K Caucasians and three other subsets contain 50K unlabeled images of non-Caucasians, respectively. We use Caucasian as source domain and other races as target domains in our experiments.

\subsection{Implementation details}

For the baseline network, we employ the widely used VGGNet \cite{simonyan2014very} and ResNet-34 \cite{he2016deep} architecture. We finetune the VGG model \cite{parkhi2015deep} with the guidance of Softmax loss on the CAISA-Webface, and is called VGG(finetune) model in our paper; while the ResNet-34 is trained with the guidance of Arcface loss \cite{deng2018arcface} on the CAISA-Webface, and is called Arcface model in our paper.

For data processing of VGG, all the images of different datasets are aligned to the same reference point using three facial landmarks (left eye, right eye and center of mouth). The images are firstly resized to $250\times250$ and are then randomly cropped to $224\times224$. We also augment the data by flipping it horizontally with 50\% probability. And for data processing of ResNet, we use five facial landmarks for similarity transformation, then crop and resize the faces to 112$\times$112. Each pixel ([0, 255]) in RGB images is normalized by subtracting 127.5 and then being divided by 128.

For training CDA(vgg-soft) model, we select VGG model \cite{parkhi2015deep} which uses VGGNet \cite{simonyan2014very} and is trained on VGGface dataset \cite{parkhi2015deep} and reports excellent results on LFW and YTF benchmarks. However, we know nothing about the face aligned method in VGG model which may cause inconsistent alignment methods between training data and test data and thus results in a poor performance. To address this issue, We use the fine-tuning architecture similar to \cite{Tzeng2014Deep,Long2015Learning} where CASIA-WebFace dataset \cite{yi2014learning} is utilized as source data to fine-tune the VGG model. The CASIA-WebFace dataset and other target datasets share the uniform alignment methods as we mentioned before. The based learning rate is fixed at $10^{-4}$. As the last classifier is trained from scratch, we set its learning rate to be 10 times that of the lower layers. The batch size is set to 32 and the network is trained for $2\times10^4$ iterations.

After fine-tuning the VGG model with our source data, we utilize the unlabeled target data and labeled source data to adapt the baseline network by MMD. Our network architecture is comprised of two basic CNNs which are identical in structure and shared by parameters. One is for classification on source data and the other is for representation learning on target data. We use Softmax loss as source classification loss and fix the learning rate of all layers to $10^{-4}$. The hyper-parameter $\lambda$ in Eq. \ref{DA} is fixed at 0.5. The kernel in MMD is Gaussian kernel $k({x^s},{x^t}) = {e^{ - {{\left\| {{x^s} - {x^t}} \right\|}^2}/\gamma }}$ where $\gamma$ donates the bandwidth. In our experiments, DAN(vgg-soft) \cite{Long2015Learning} applies multi-kernel MMD on both $fc6$ and $fc7$ layer. Five Gaussian kernels are utilized by setting bandwidth to ${\gamma}_m\cdot(1,2^1,2^2,2^3,2^4)$ where ${\gamma}_m$ is set to the median pairwise distances \cite{gretton2012optimal} on training data. DDC(vgg-soft) \cite{Tzeng2014Deep} adopts single-kernel MMD on $fc7$ layer, and it only utilizes one Gaussian kernel in which bandwidth is set to ${\gamma}_m$. To evaluate the effectiveness of multi-layer and multi-kernel adaptation more comprehensively, we further make several variants of MMD-based network, namely single-kernel MMD on both $fc6$ and $fc7$ layer and multi-kernel MMD on $fc7$ layer. We denote them as $DDC_{ml}$(vgg-soft) and $DDC_{mk}$(vgg-soft), respectively.

For our clustering methods, the hyper-parameter $p$ is set to be 3. We set the parameter $\alpha$ and $\beta$ in Eq. \ref{cluster} and Eq. \ref{scattered} as 0.675 and 0.8 in CASIA$\rightarrow$GBU task, and set them as 0.65 and 0.8 in CASIA$\rightarrow$IJB-A/IJB-B/IJB-C task. After obtaining the pseudo-labels, we further fine-tune the target network with them. We use Softmax loss as target pseudo classification loss. The learning rate is started from $1e-4$ and decreased twice with a factor of 10 when errors plateau. The network is trained for $2\times10^4$ iterations. We set the batch size, momentum, and weight decay as 64, 0.9 and $5e-4$, respectively.

For training CDA(res-arc) model, we first train a Arcface model with the guidance of Arcface loss \cite{deng2018arcface} on the CAISA-Webface. We set the batch size, momentum, and weight decay as 200, 0.9 and 5e-4, respectively. The learning rate is started from 0.1 and decreased twice with a factor of 10 when errors plateau. After that, we utilize the unlabeled target data and labeled source data to adapt Arcface model by MMD. We use Arcface loss as source classification loss and fix the learning rate of all layers to 1e-3. The hyper-parameter $\lambda$ in Eq. \ref{DA} is fixed at 5. DAN(res-arc) \cite{Long2015Learning} applies multi-kernel MMD on last two fully-connected layers. For our clustering methods, we set the parameter $\alpha$ and $\beta$ in Eq. \ref{cluster} and Eq. \ref{scattered} as 0.8 and 0.85 in CASIA$\rightarrow$GBU task, and set them as 0.7 and 0.85 in CASIA$\rightarrow$IJB-A/IJB-B/IJB-C task. After obtaining the pseudo-labels, we further fine-tune the target network with them. We use Softmax loss as target pseudo classification loss. The learning rate is 1e-3. We set the batch size, momentum, and weight decay as 200, 0.9 and 5e-4, respectively. Other experimental settings are similar to CDA(vgg-soft).

\begin{table}[htbp]
\small
\caption{VR at FAR of 0.001 for GBU partitions \cite{Phillips2012The}. }
\begin{center}
\begin{threeparttable}
\setlength{\tabcolsep}{3mm}{
\begin{tabular}{l|ccc}
\hline
\textbf{Method} & \textbf{Ugly} & \textbf{Bad} & \textbf{Good}\\
\hline\hline
LRPCA-face \cite{Phillips2012The} & 7.00\% & 24.00\% & 64.0\% \\
Fusion \cite{phillips2017cross} & 15.00\% & 80.00\% & 98.00\% \\
VGG \cite{phillips2017cross} & 26.00\% & 52.00\% & 85.00\% \\
Arcface\tnote{1} \cite{deng2018arcface} & 75.00\% & 90.32\% & 96.21 \% \\
\hline
VGG(finetune)\tnote{2} & 48.80\% & 73.55\% & 95.57\% \\
$DDC$(vgg-soft)\tnote{3} \cite{Tzeng2014Deep} & 60.90\% & 86.68\% & 98.24\%  \\
$DDC_{ml}$(vgg-soft)\tnote{3} & 63.42\% & 87.08\% & 98.54\% \\
$DDC_{mk}$(vgg-soft)\tnote{3} & 68.42\% & 87.68\% & 98.67\%  \\
$DAN$(vgg-soft)\tnote{3} \cite{Long2015Learning}& 69.42\% & 88.87\% & 98.93\%  \\ \hline
CDA(vgg-soft) (ours) & 73.58\% & 92.93\% & \textbf{99.18\%} \\
\textbf{CDA(res-arc) (ours)} & \textbf{83.96\%} & \textbf{94.84\%} & 97.81\% \\
\hline
\end{tabular}}
 \begin{tablenotes}
  \item[1] Arcface is one of our baseline networks. It uses ResNet-34 architecture and is trained with the guidance of Arcface loss \cite{deng2018arcface} on the CAISA-Webface.
  \item[2] VGG(finetune) is one of our baseline networks. It finetunes the VGG model \cite{parkhi2015deep} supervised with Softmax on CASIA-WebFace dataset.
  \item[3] $DDC$, $DDC_{ml}$, $DDC_{mk}$ and $DAN$ represent the variants of MMD-based network.
\end{tablenotes}
\end{threeparttable}
\end{center}
\label{tab1}
\end{table}

\subsection{Experiment Results}

\begin{figure*}
\centering
\subfigure[GBU (Good)]{
\label{fig5a} 
\includegraphics[width=5.4cm]{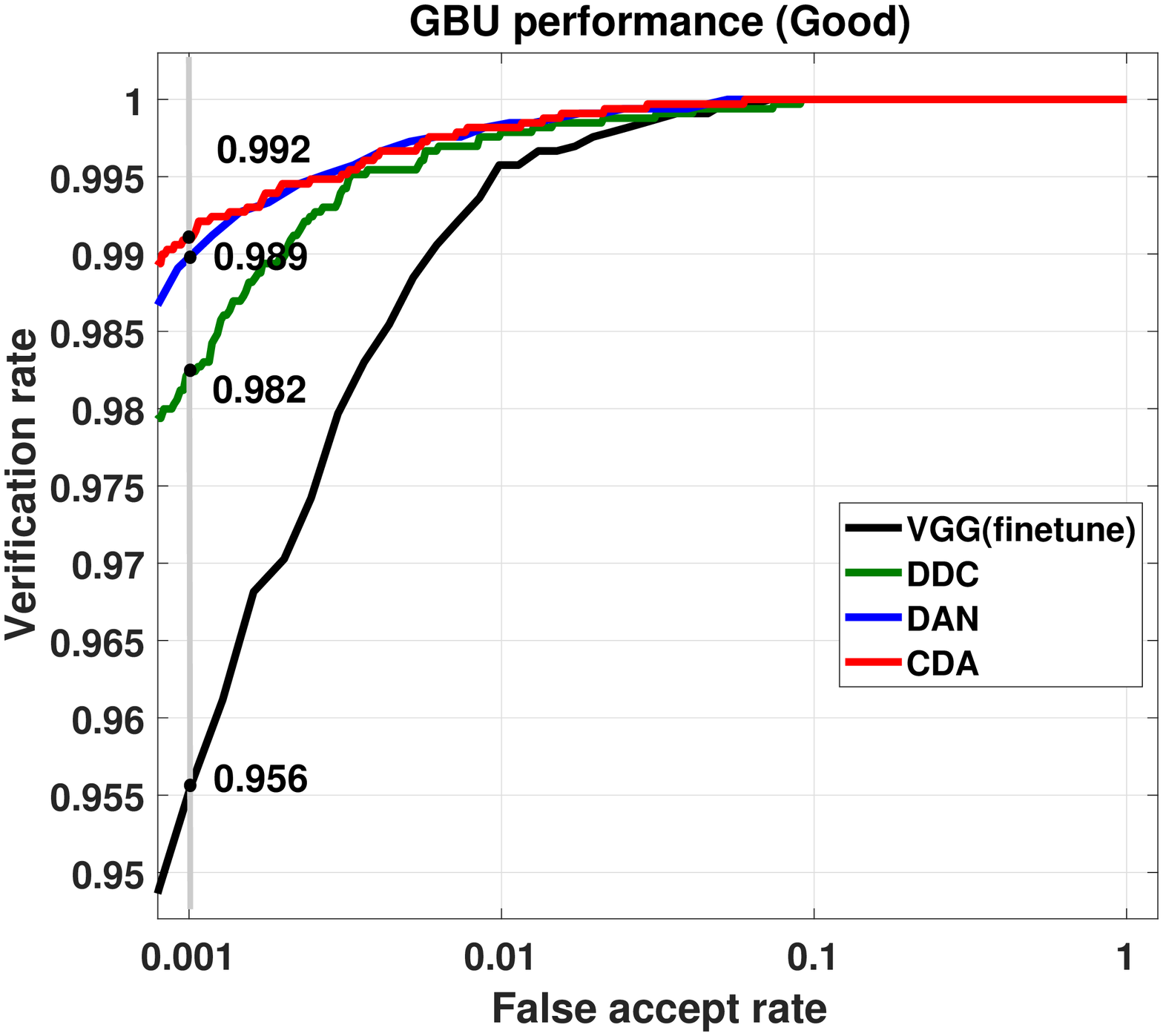}}
\hspace{0cm}
\subfigure[GBU (Bad)]{
\label{fig5b} 
\includegraphics[width=5.4cm]{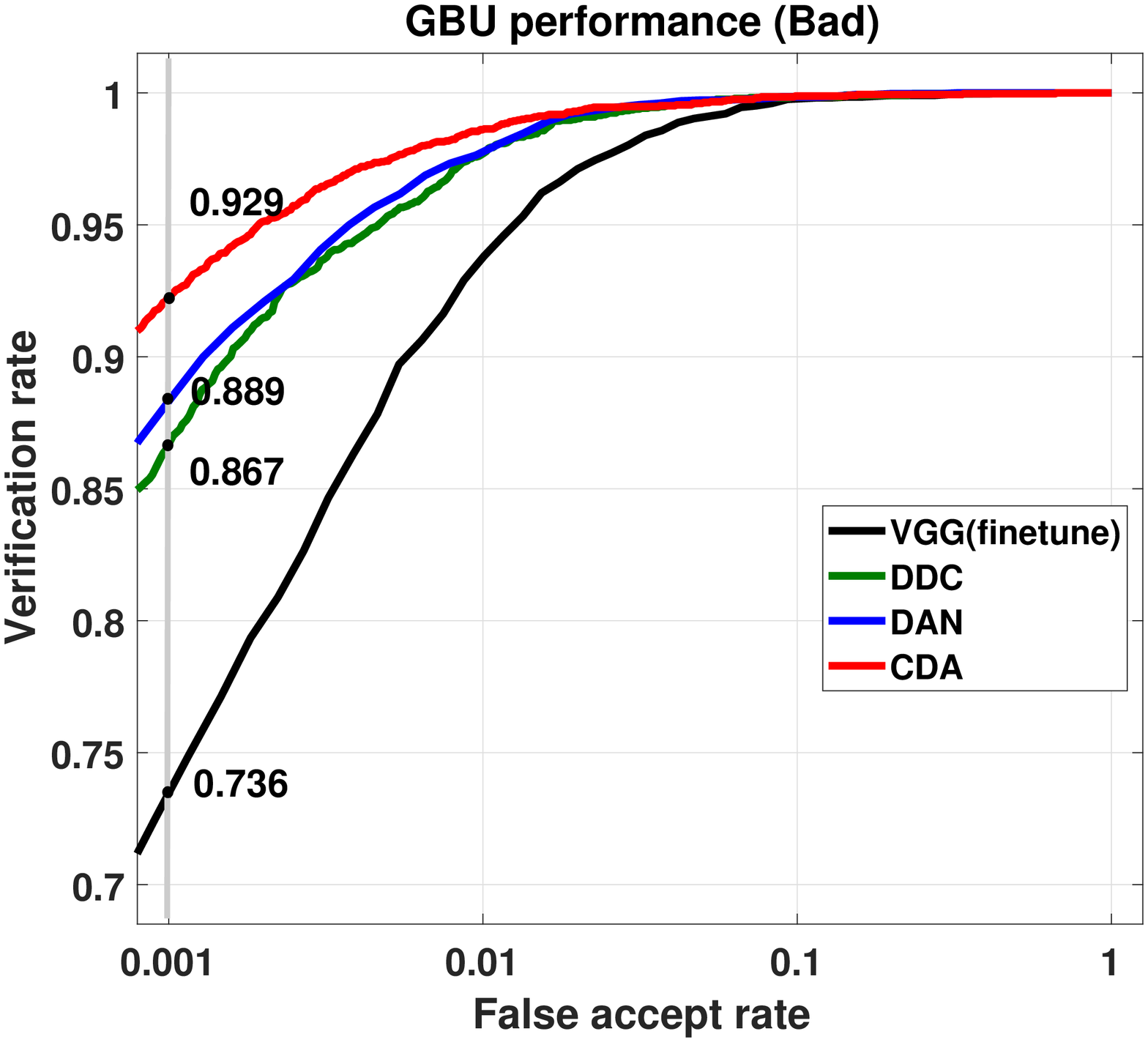}}
\subfigure[GBU (Ugly)]{
\label{fig5c} 
\includegraphics[width=5.4cm]{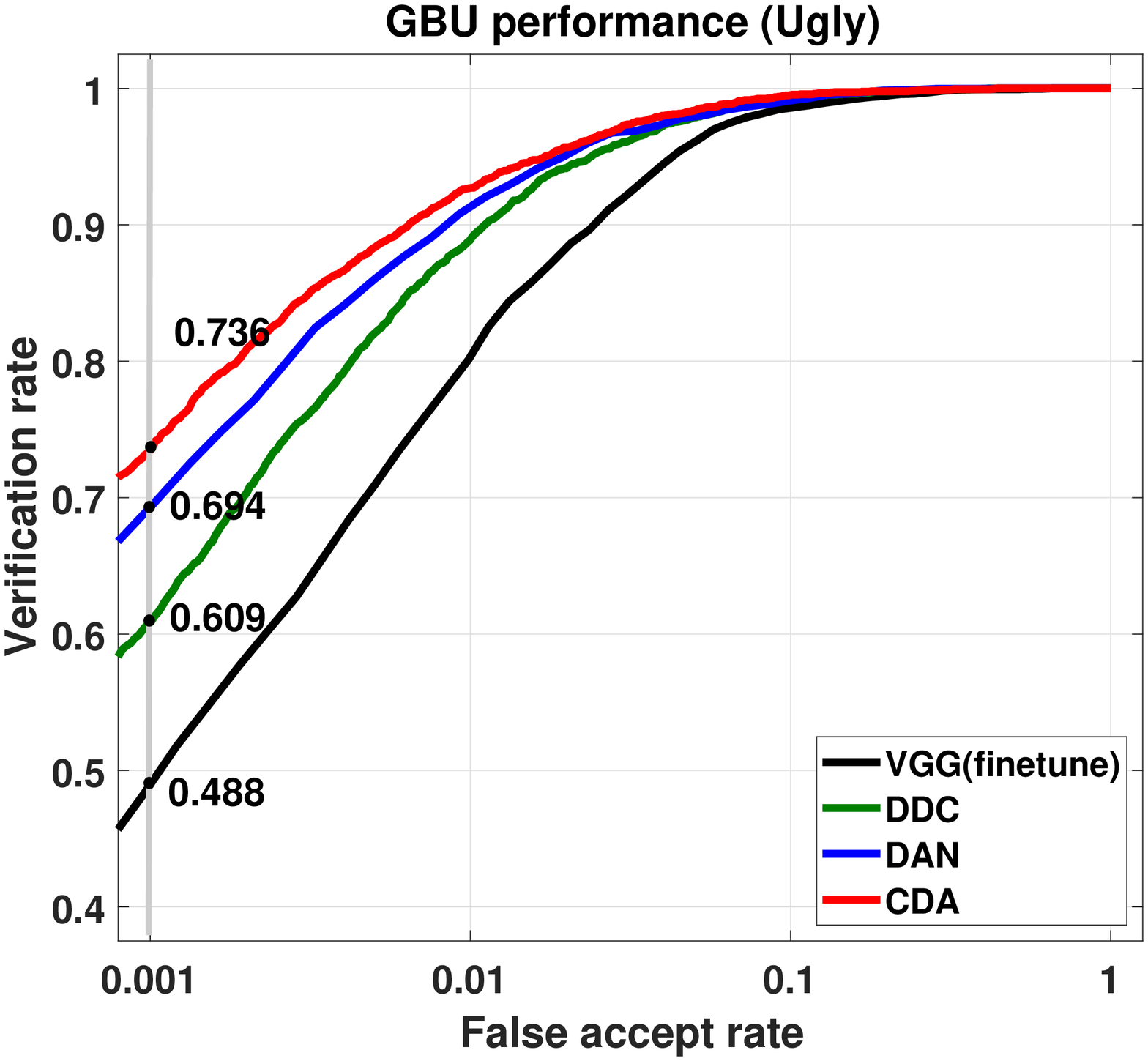}}
\caption{The ROC curves on (a) Good, (b) Bad and (c) Ugly partition of GBU database. Black lines are ROC curves of \emph{VGG(finetune)} model; Green lines are those of \emph{DDC(vgg-soft)} model; Blue lines are those of \emph{DAN(vgg-soft)} model; Red lines are those of our \emph{CDA(vgg-soft)} model. The verification rate for each partition at a FAR of 0.001 is highlighted by the vertical lines at FAR=0.001.}
\label{fig5} 
\end{figure*}

\textbf{CASIA$\rightarrow$GBU.} In the experiment of GBU dataset \cite{Phillips2012The}, we report the verification rate at a FAR of 0.001 and ROC curve for three partitions, i.e. the Good, the Bad and the Ugly. \emph{Fusion} method in \cite{phillips2017cross} denotes the FRVT 2006 fusion algorithm and the result \emph{VGG} was reported in \cite{phillips2017cross} by utilizing the VGG model \cite{parkhi2015deep}. The \emph{LRPCA-face} model is a baseline algorithm in GBU dataset \cite{Phillips2012The} which is a refined implementation of the standard PCA-based FR algorithm. The \emph{VGG(finetune)} represents one of our baseline networks which finetunes the VGG model with CASIA-WebFace dataset \cite{yi2014learning}; and Arcface is the other baseline network which uses ResNet-34 architecture and is trained with the guidance of Arcface loss \cite{deng2018arcface} on the CAISA-Webface. The exact results are shown in Table \ref{tab1} and Fig. \ref{fig5}.

From the results, we can see several important observations. \textbf{(1)} For Ugly partition, all the models give the accuracies of less than 84\% and specially an extremely low total accuracy of 15\% with \emph{Fusion} model, showing face verification on Ugly partition is a very challenging task despite of its frontal faces. Significantly, the performance of deep models is unsatisfactory as well and \emph{VGG} only achieves 26\% on Ugly partition, which illustrates the limitation of existing deep models trained with Web-collected dataset and the necessity of adopting UDA in FR tasks. \textbf{(2)} Compared with the results of \emph{VGG} reported in \cite{phillips2017cross}, our baseline model fine-tuned \emph{VGG} with CASIA-WebFace \cite{yi2014learning} obtains much better performance, which improves the accuracy to 48.80\%, 73.55\%, 95.57\% on Ugly, Bad and Good partition. The results suggest that the uniform face aligned algorithm of training and testing data is the key to ensure performance in the FR problem. \textbf{(3)} MMD-based networks, i.e. $DDC$(vgg-soft)\cite{Tzeng2014Deep}, $DDC_{ml}$(vgg-soft), $DDC_{mk}$(vgg-soft)and $DAN$(vgg-soft) \cite{Long2015Learning}, substantially outperform \emph{VGG(finetune)} model on target dataset. This confirms that incorporating MMD to deep networks and minimizing the domain discrepancy are really helpful. \textbf{(4)} Single-kernel MMD models ($DDC$(vgg-soft) and $DDC_{ml}$(vgg-soft)) obtain a little bit worse results compared with multi-kernel MMD ($DDC_{mk}$(vgg-soft) and $DAN$(vgg-soft)). It is because multiple kernels with different bandwidths can match both the low-order moments and high-order moments resulting in a better alignment of distribution of source and target domain. \textbf{(5)} The \emph{DAN(vgg-soft)} obtains the best performances compared with other MMD-based networks, which superior to our \emph{VGG(finetune)} by about 20.62\% on the Ugly, 14.13\% on the Bad and 3.1\% on the Good. In addition to multi-kernel adaptation, \emph{DAN(vgg-soft)} is also benefited from multi-layer adaptation. In deep networks, representations of different layers correspond to different levels of abstraction, changing from low-level primary elements to multifarious facial attributes. Hence the hidden representations of all the task-specific layers need to be matched to consolidate the adaptation quality at all levels. \textbf{(6)} When introducing clustering algorithms and pseudo-labels into \emph{DAN(vgg-soft)} models, the performances of our \emph{CDA(vgg-soft)} method further improve and obtain the best performances with 73.58\%, 92.93\% and 99.18\% for Ugly, Bad and Good set. We can draw conclusions that only aligning the feature space through MMD is not enough for FR and that further learning target discriminative representations using pseudo-labels is an effective way to boost the performance. Moreover, the results quantificationally prove the good quality of pseudo-labels generated by our clustering method. \textbf{(7)} Without adaptation, Arcface  \cite{deng2018arcface}, which published in CVPR'19 and reported SOTA performance on the LFW and MegaFace challenges, can not obtain perfect performance on GBU due to domain gap. Our CDA(res-arc) can outperform Arcface method and even achieve about 3\% gains on Ugly partition.

\begin{figure*}
\centering
\subfigure[ROC curve]{
\label{fig12a} 
\includegraphics[width=5.3cm]{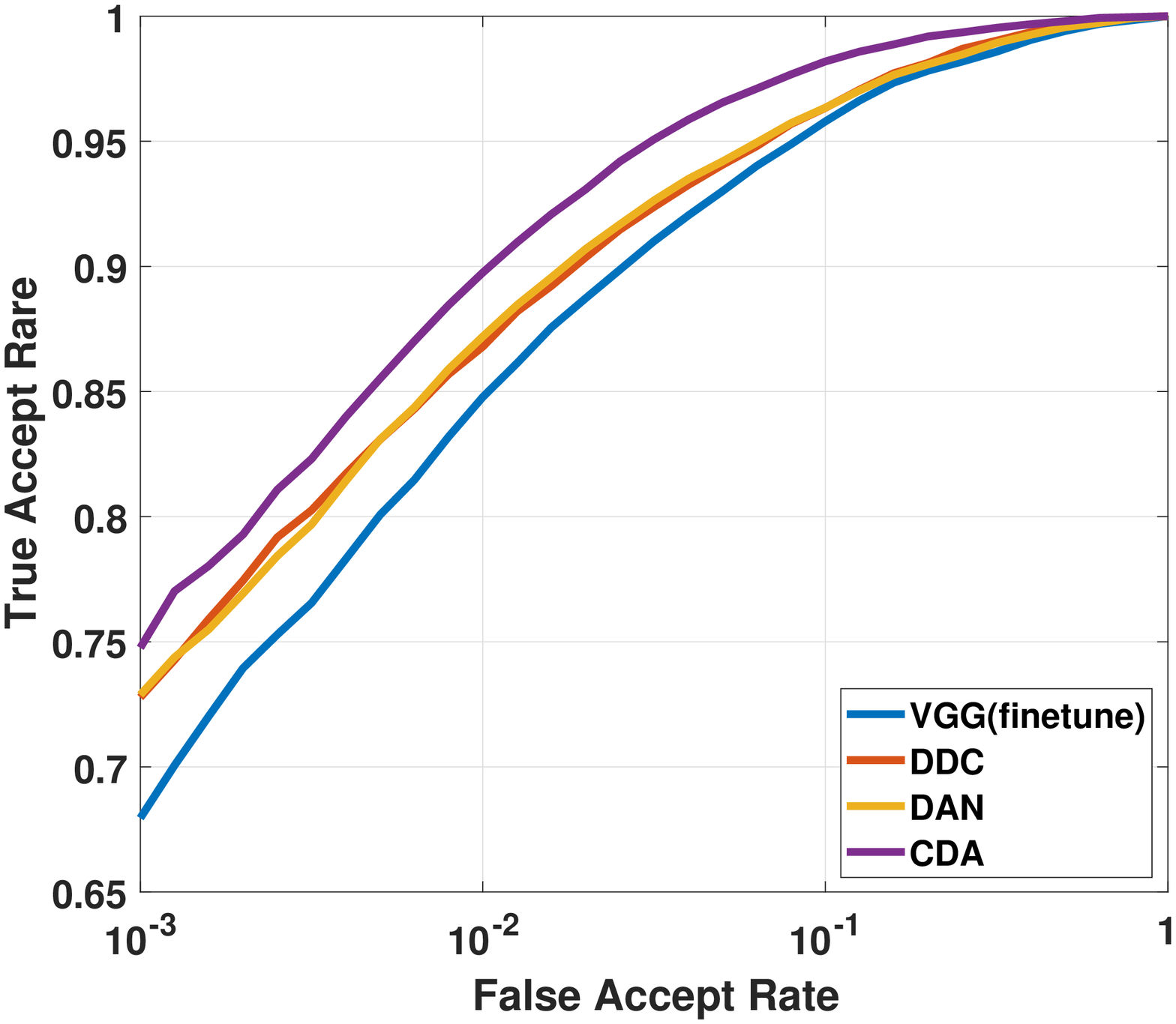}}
\hspace{0cm}
\subfigure[DET curve]{
\label{fig12b} 
\includegraphics[width=5.3cm]{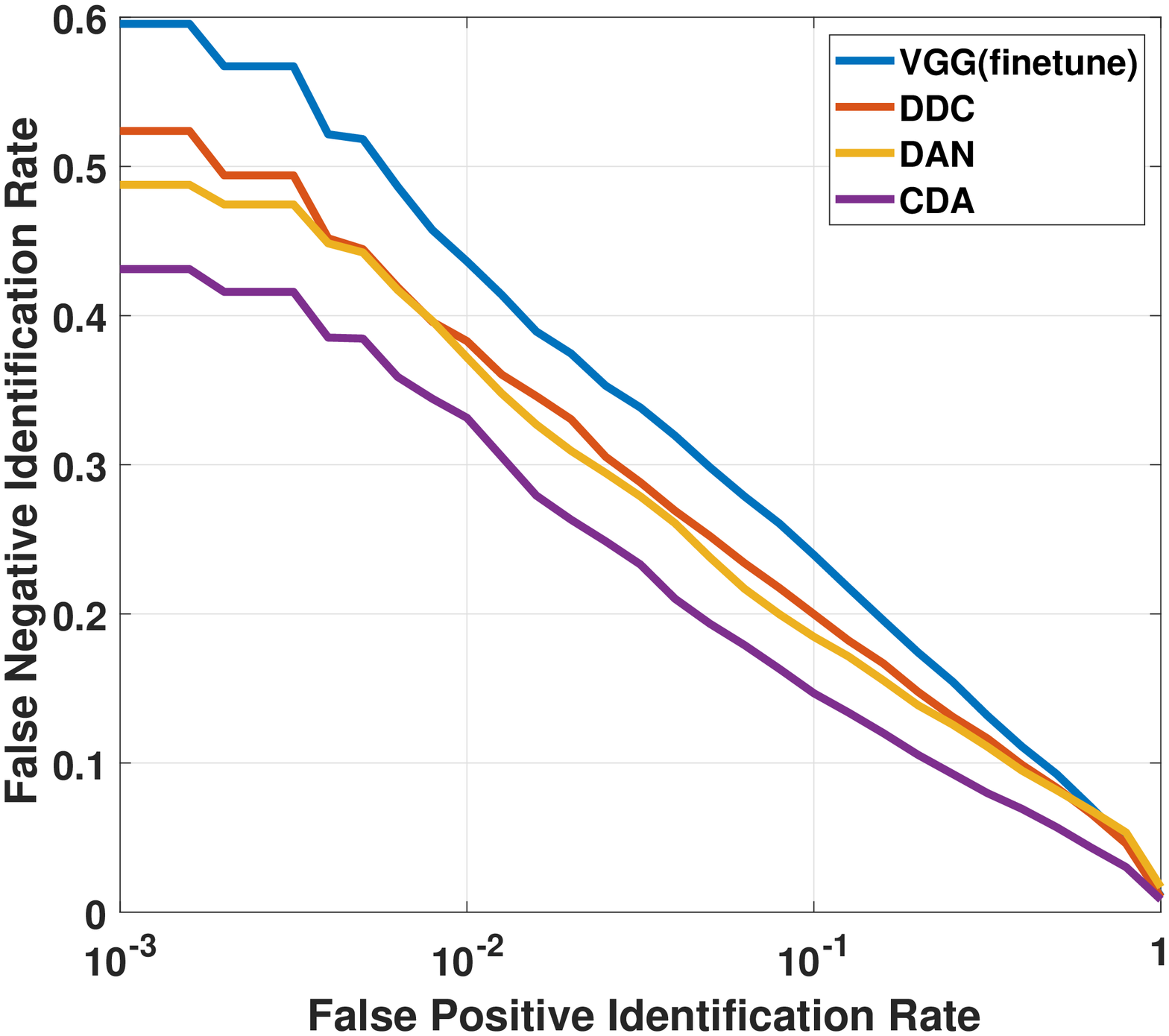}}
\subfigure[CMC curve]{
\label{fig12c} 
\includegraphics[width=5.3cm]{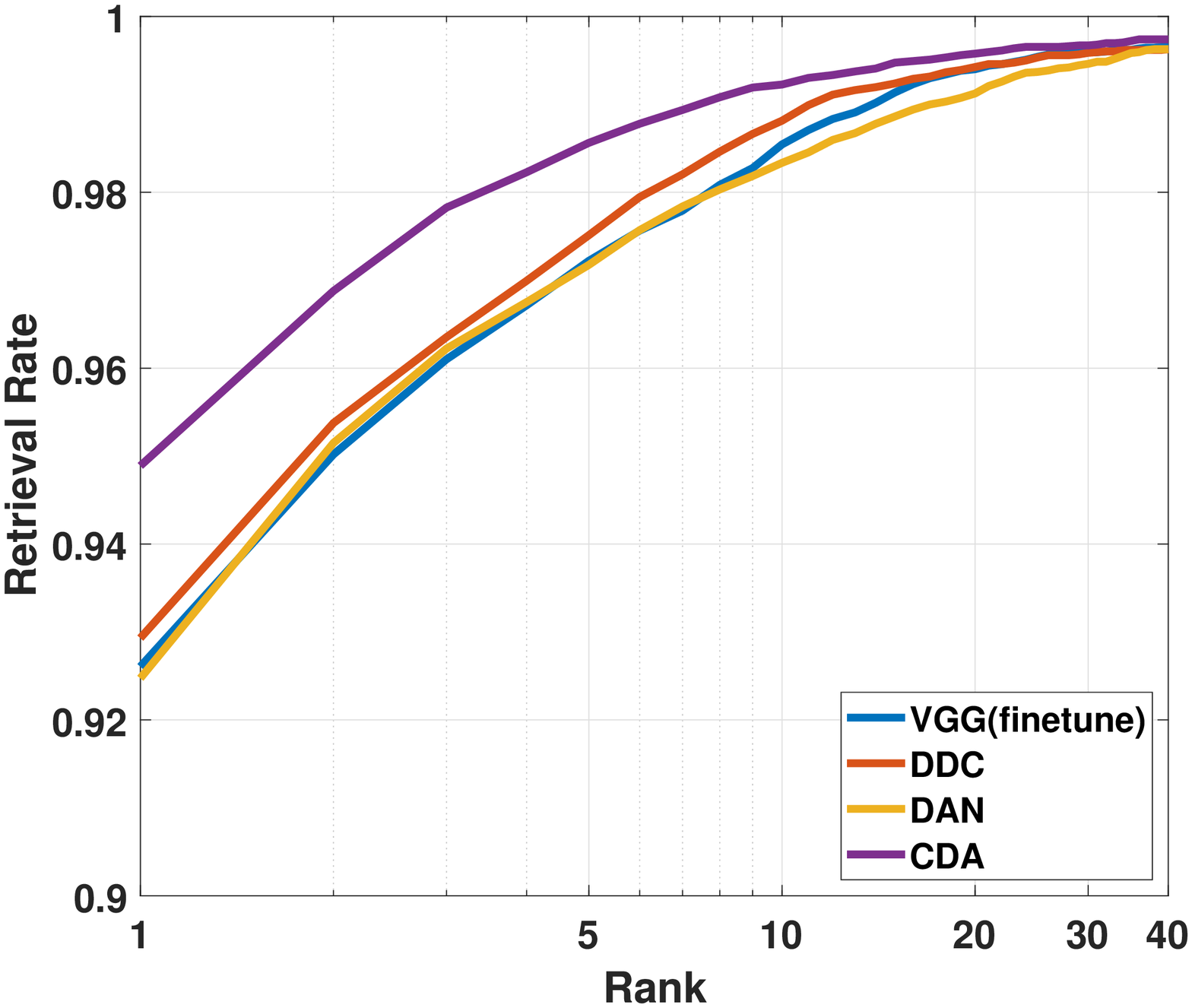}}
\caption{Results on the IJB-A dataset (average over 10 splits). (a) ROC curve for the compare protocol (higher is better). (b) DET curve for the search protocol (lower is better). (c) CMC curve for the search protocol (higher is better).}
\label{fig12} 
\end{figure*}

\textbf{CASIA$\rightarrow$IJB-A.} We perform experiments in two settings on the IJB-A benchmark dataset \cite{klare2015pushing}: the TAR at different FAR of 0.1, 0.01, and 0.001 for verification; the TPIR at different FPIR of 0.1, 0.01 and the rank-1, rank-10 accuracy for identification. Table \ref{tab4} and Fig. \ref{fig12} report the results of face verification and identification. We can observe that the \emph{VGG} model does not perform well on IJB-A benchmarks. Benefiting from the same aligned method of training and testing data, \emph{VGG(finetune)} model obtains a little promotion compared to \emph{VGG} model, but its performance is still imperfect. The images and video frames in IJB-A dataset \cite{klare2015pushing} contains full pose variation and a wide variation in imaging conditions and geographic origin. It is challenging for models trained with VGGface database \cite{parkhi2015deep} or CASIA-Webface databases \cite{yi2014learning} due to large domain gap. For example, video frames in IJB-A database are likely to be degraded for motion or out-of-focus blur, compression noise or scale variations. When we reduce their domain gap using MMD-based networks, the improvement
becomes more significant. Especially, \emph{DAN(vgg-soft)} boosts around 9\% TAR@FAR=0.001 for verification, and around 15\% FNIR@FPIR=0.01 for identification compared with \emph{VGG} model. It proves that the source networks trained with frontal and high-definition faces can adapt to recognize the blur images of large pose variations to a certain extent through domain adaptation. Similar to the experiments on GBU, multi-layer MMD also attains higher accuracy than single-layer MMD in most cases, which confirms the capability of multi-layers for distribution adaptation. 
After introducing clustering algorithms and pseudo-labels into \emph{DAN(vgg-soft)}, the \emph{CDA(vgg-soft)} model surpasses other methods and outperforms \emph{DAN(vgg-soft)} by about 2-4\% on all metrics, which further demonstrates the advantage of our clustering algorithms. Further, when compared with the SOTA methods, i.e. Arcface,  our \emph{CDA(res-arc)} can still obtain better performance.

\begin{table*}[htbp]
\begin{center}
\caption{Performance evaluation on the IJB-A dataset \cite{klare2015pushing}. The results are averaged over 10 testing splits.}
\begin{threeparttable}
\setlength{\tabcolsep}{3mm}{
\begin{tabular}{l|ccc|ccccc}
\hline
\multirow{2}{*}{\textbf{Method}} & \multicolumn{3}{c|}{\textbf{IJB-A Verification TAR}} & \multicolumn{4}{c}{\textbf{IJB-A Identification TPIR}}\\
&FAR=0.001 & FAR=0.01 & FAR=0.1 & FPIR=0.01 & FPIR=0.1 & Rank-1 &  Rank-10 \\\hline
Bilinear-CNN \cite{chowdhury2016one} & - & - & - & 14.20\% & 34.10\% & 58.80\% & - \\
Face-Search \cite{wang2015face} & - & 73.30\% & - & 38.30\% & 61.30\% & 82.00\% & - \\
Deep-Multipose \cite{abdalmageed2016face} & - & 78.70\% & - & 52.00\% & 75.00\% & 84.60\% & 94.70\% \\
Triplet-Similarity \cite{sankaranarayanan2016triplet} & - & 79.00\% & - & 55.60\% & 75.41\% & 88.01\% & 97.38\% \\
Joint Bayesian \cite{chen2016unconstrained} & - & 83.80\% & - & 57.68\% & 78.97\% & 90.30\% & 97.70\% \\
VGG \cite{parkhi2015deep} & 64.19\%  &  84.02\%  &  96.09\%  &  47.37\%  &  74.30\%  &  91.11\%  &  98.25\% \\
Arcface\tnote{1} \cite{deng2018arcface} & 74.19\% & 87.11\% & 94.87\% & 65.36\%  &  80.71\% & 90.68\% & 96.07\% \\ \hline
VGG(finetune)\tnote{2}& 67.96\%  &  84.78\%  &  95.80\%  &  56.36\%  &  76.05\%  &  92.61\%  &  98.54\%  \\
$DDC$(vgg-soft) \cite{Tzeng2014Deep} & 72.78\%  &  86.80\%  &  96.34\%  &  61.71\%  &  80.02\%  &  92.93\%  &  98.81\% \\
$DDC_{ml}$(vgg-soft)\tnote{3}& 72.97\%  &  87.74\%  &  96.70\%  &  62.82\%  &  81.30\%  &  92.91\%  &  98.62\%  \\
$DDC_{mk}$(vgg-soft)\tnote{3}& 72.53\%  & 87.13\%  &  96.54\%  &  61.58\%  &  82.33\%  &  92.54\%  &  98.52\%  \\
$DAN$(vgg-soft) \cite{Long2015Learning} & 72.88\%  &  87.20\%  &  96.34\%  &  62.81\%  &  81.54\%  &  92.47\%  &  98.33\% \\ \hline
CDA(vgg-soft)(ours)  & 74.76\%  &  89.76\%  &  \textbf{98.19\%}  &  66.85\%  &  85.32\%  &  \textbf{94.89\%}  &  \textbf{99.23\%} \\
\textbf{CDA(res-arc) (ours)} &  \textbf{82.45\%} &  \textbf{91.11\%} &  96.96\% & \textbf{75.49\%} & \textbf{87.76\%} &  93.61\% &  97.62\% \\
\hline
\end{tabular}}
 \begin{tablenotes}
  \item[1] Arcface is one of our baseline networks. It uses ResNet-34 architecture and is trained with the guidance of Arcface loss \cite{deng2018arcface} on the CAISA-Webface.
  \item[2] VGG(finetune) is one of our baseline networks. It finetunes the VGG model \cite{parkhi2015deep} supervised with Softmax on CASIA-WebFace dataset.
  \item[3] $DDC_{ml}$ adopts single-kernel MMD on both $fc6$ and $fc7$ layer and $DDC_{mk}$ adopts multi-kernel MMD on $fc7$ layer.
\end{tablenotes}
\end{threeparttable}
\end{center}
\label{tab4}
\end{table*}

\begin{table*}[htbp]
\caption{Performance evaluation on the IJB-B \cite{whitelam2017iarpa} and IJB-C \cite{maze2018iarpa} dataset.}
\begin{center}
\begin{threeparttable}
\setlength{\tabcolsep}{1.5mm}{
\begin{tabular}{l|ccc|cc|ccc|cc}
\hline
\multirow{3}{*}{\textbf{Method}} & \multicolumn{5}{c|}{\textbf{IJB-B}} & \multicolumn{5}{c}{\textbf{IJB-C}} \\ \cline{2-11}
& \multicolumn{3}{c|}{\textbf{ Verification TAR@FAR}} & \multicolumn{2}{c|}{\textbf{ Identification}} & \multicolumn{3}{c|}{\textbf{ Verification TAR@FAR}} & \multicolumn{2}{c}{\textbf{ Identification}}\\
&0.001 & 0.01 & 0.1  & Rank-1 &  Rank-10 & 0.001 & 0.01 & 0.1  & Rank-1 &  Rank-10 \\\hline
GOTS-1 \cite{whitelam2017iarpa} & 33.00\% & 60.00\% & 78.00\% & 42.00\%  & 62.00\% &  - & - & - & - & - \\
GOTS-2 \cite{maze2018iarpa}& - & - & - & - & - & 32.00\% & 62.00\% & 80.00\% & -   & - \\
FaceNet \cite{schroff2015facenet}& - & - & - & - & - & 66.00\% & 82.00\% & 92.00\% & - & - \\
DR-GAN \cite{tran2017disentangled} & - & - & - & - & - & 66.10\% & 82.40\% & - & 70.80\% &  82.80\% \\
VGG \cite{parkhi2015deep} & 72.00\% & 86.00\% & - & 78.00\% &  89.00\% & 75.00\%  &  86.00\%  &  95.00\%  &   -  &  - \\
Bodla et al. \cite{bodla2017deep} & 83.00\%  &   92.50\%  &  -  &  -   &  - &  - & - & - & - & - \\
Yin et al. \cite{yin2019towards} & - & - & - & - & - & 75.60\% & 89.20\% & - & 77.60\%  & 86.10\% \\
Arcface\tnote{1} \cite{deng2018arcface}& 86.11\%  &  93.40\%  &  97.66\%  &  \textbf{86.43\%}  &  93.33\% & \textbf{88.88\%}  &  94.76\%  &  98.10\%  &  88.05\%    & 93.56\%  \\ \hline
\textbf{CDA(res-arc) (ours)} & \textbf{87.35\%}  &  \textbf{94.55\%}  &  \textbf{98.08\%}  &  86.22\%   &  \textbf{93.33\%}  & 88.06\%  &  \textbf{94.85\%}  &  \textbf{98.33\%}  &  \textbf{88.19\%}   &  \textbf{93.70\%} \\
\hline
\end{tabular}}
 \begin{tablenotes}
  \item[1] Arcface here is our baseline network which uses ResNet-34 architecture and is trained with the guidance of Arcface loss \cite{deng2018arcface} on the CAISA-Webface.
\end{tablenotes}
\end{threeparttable}
\end{center}
\label{IJBC}
\end{table*}

\textbf{CASIA$\rightarrow$IJB-B/C.} We perform experiments in two settings on the IJB-B and IJB-C benchmark dataset \cite{whitelam2017iarpa,maze2018iarpa}: the TAR at different FAR of 0.1, 0.01, and 0.001 for verification; the rank-1 and rank-10 accuracy for identification. Table \ref{IJBC} reports the results of face verification and identification. We compare our proposed method with Government-off-the-shelf (GOTS-1 \cite{whitelam2017iarpa}), Bodla et al. \cite{bodla2017deep}, VGG \cite{parkhi2015deep} and Arcface \cite{deng2018arcface} on IJB-B dataset; and compare our method with GOTS-2 \cite{maze2018iarpa}, FaceNet \cite{schroff2015facenet}, DR-GAN \cite{tran2017disentangled}, Yin et al. \cite{yin2019towards}, VGG \cite{parkhi2015deep} and Arcface \cite{deng2018arcface} on IJB-C dataset. From the results, we can see that our \emph{CDA(res-arc)} achieves improvement over the previous SOTA methods, i.e. Arcface, with TAR of 87.35\% at FAR = 10e-3 on IJB-B; while on IJB-C, it achieves a Rank1 accuracy of 88.19\% in face identification. In our \emph{CDA}, MMD-based networks address huge domain discrepancy to learn transferable representations and provide more reliable underlying face representation for clustering; while pseudo-labels further learn more discriminative representations for FR tasks. Actually, in our experiments, we just utilized limited number of images in IJB-A as target training data to achieve such improvement on these two challenging benchmarks. If more target training data are used to adapt source model, more significant improvement can be obtained.

\textbf{Caucasian$\rightarrow$Non-Caucasian.} Some papers \cite{wang2019racial,wang2019mitigate} have proved that existing face recognition algorithms indeed suffer from racial bias. Due to the domain gap among different races, training and testing on different races results in severe performance drop. To validate the effectiveness of our domain adaptation method, we adopt CDA to transfer knowledge among different races. We use BUPT-Transferface as training data, and use RFW \cite{wang2019racial} as testing data. Labeled Caucasians are utilized as source domain and unlabeled Indians/Asians/Africans are utilized as target domains in our experiments. The results are given in Table \ref{race} and we have the following observations. \textbf{(1)} The Softmax and Arcface model which are trained on Caucasians perform well on Caucasian testing subset, but the accuracy drops on Asian and African because of domain gap. For example, the accuracy of the ArcFace model on Caucasian testing subset reaches 94.78\%, but its accuracy dramatically decreases to less than 85.13\% on Asian subset. \textbf{(2)} \emph{DDC}(res-soft) \cite{Tzeng2014Deep} and \emph{DAN}(res-soft) \cite{Long2015Learning} align Caucasian domain and other race domain with help of MMD. But they are only superior to baseline by about 1-2\% which confirms our thought that only aligning domains globally is not enough for face recognition. \textbf{(3)} When adopting clustering algorithms and pseudo-labels, our \emph{CDA(res-soft)} and \emph{CDA(res-arc)} model outperform the baseline models, especially \emph{CDA(res-arc)} obtains the best performances with 92.08\%, 88.80\% and 88.12\% on Indian, Asian and African set.

\begin{table}
\caption{Verification accuracy (\%) on 6000 pairs of RFW dataset \cite{wang2019racial}. ``(res-soft)" represents the ResNet-34 methods using Softmax as source classification loss; while ``(res-arc)" represents the ones using Arcface.}
	\begin{center}
    \small
    \begin{threeparttable}
    \setlength{\tabcolsep}{1.5mm}{
	\begin{tabular}{c|cccc}
		\hline
         Methods & Caucasian & Indian & Asian & African \\ \hline \hline
         Softmax\tnote{1} & 94.12\% & 88.33\% & 84.60\% & 83.47\% \\
         $DDC$(res-soft) \cite{Tzeng2014Deep}& - & 90.53\% & 86.32\% & 84.95\% \\
         $DAN$(res-soft \cite{Long2015Learning} & - & 89.98\% &85.53\% & 84.10\% \\
         \textbf{CDA(res-soft) (ours)} & - & \textbf{90.73\%} & \textbf{88.88\%}&  \textbf{87.42\%}\\ \hline
         Arcface\tnote{1} \cite{deng2018arcface} & 94.78\% &90.48\% & 86.27\% & 85.13\%  \\
         $DDC$(res-arc) \cite{Tzeng2014Deep}  & - &91.63\% &87.55\% & 86.28\% \\
         $DAN$(res-arc) \cite{Long2015Learning} & - & 91.78\% & 87.78\% & 86.30\%  \\
         \textbf{CDA(res-arc) (ours)} & - & \textbf{92.08\%} & \textbf{88.80\%} & \textbf{88.12\%} \\ \hline
	\end{tabular}}
	 \begin{tablenotes}
  \item[1] Softmax and Arcface here are our baseline networks which use ResNet-34 architecture trained on the CAISA-Webface.
\end{tablenotes}
\end{threeparttable}
    \end{center}
    \label{race}
\end{table}

\subsection{Empirical analysis}

\begin{figure}
\centering
\includegraphics[width=8cm]{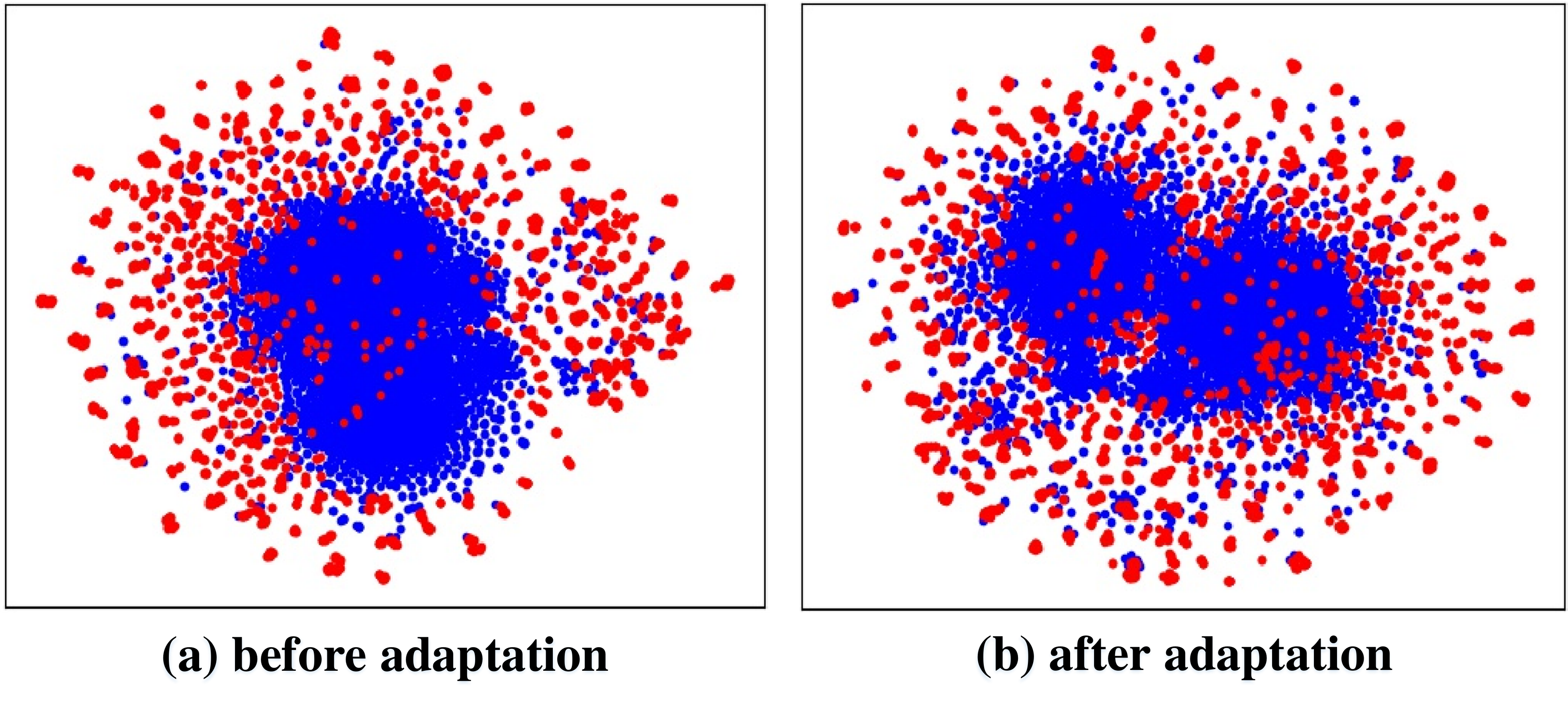}
\caption{ Feature visualization. We confirm the effects of MMD through a visualization of the learned representations using t-distributed stochastic neighbor embedding (t-SNE) \cite{maaten2008visualizing}. Blue points are source samples and red are target samples. (a) are trained without any adaptation, (b) are trained with MMD method. As we can see, compared to non-adapted method, MMD method can help our \emph{CDA} to align the source features and target features to a certain extent and improve the performance of target domain.}
\label{fig11}
\end{figure}

\textbf{Feature visualization.} To demonstrate the transferability of the MMD learned features, the visualization comparisons are conducted at feature level. First, we randomly extract the deep features of 5000 source and 5000 target images in task CASIA$\rightarrow$GBU (Ugly) with \emph{VGG(finetune)} model and \emph{DAN(vgg-soft)} model, respectively. The features are visualized using t-distributed stochastic neighbor embedding (t-SNE) \cite{maaten2008visualizing}, as shown in Fig. \ref{fig11}. Fig. \ref{fig11}(a) shows the representations without any adapt. As we can see, the distributions are separated between domains, which visually proves that there is domain gap between images of CASIA-Webface \cite{yi2014learning} and GBU database \cite{Phillips2012The}. Fig. \ref{fig11}(b) shows the result for \emph{DAN(vgg-soft)} method where features are aligned to some extent. More source and target data begin to mix in feature space so that there is not a clear boundary between them. 
Therefore, we conclude that the MMD does help our \emph{CDA(vgg-soft)} to minimize domain discrepancy and align feature space between source and target domain so that the performance of target domain improves. However, due to the particularity of face data, e.g. a larger number of identities as well as non-overlapping identities of source and target domain, misalignment still exists even after adaptation. It also verifies that MMD-adaptation is not enough for face recognition.

\begin{figure}
\centering
\includegraphics[width=5.3cm]{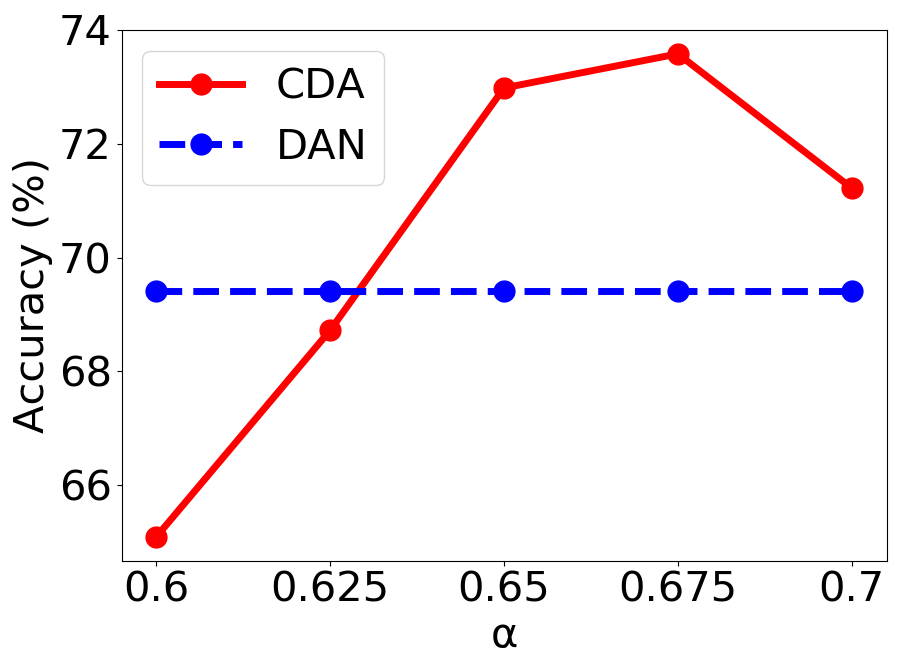}
\caption{ Parameter sensitivity of $\alpha$ (dashed lines show best DAN(vgg-soft) results).}
\label{fig10}
\end{figure}

\textbf{Parameter Sensitivity.} Besides the MMD penalty parameter $\lambda$, our clustering method involves another vital parameter $\alpha$ in Eqn. (\ref{cluster}) which controls the connection of edges in graph. Two target nodes will be connected to each other in our graph only if their cosine-similarity is larger than $\alpha$. To have a closer look at this parameter, we perform sensitivity analysis for it in transfer tasks CASIA$\rightarrow$GBU (Ugly) by varying the parameter of interest in \{0.6, 0.625, 0.65, 0.675, 0.7\}. We generate different target pseudo-labels according to different parameter $\alpha$, then fine-tune the target CNN with them respectively. The fine-tuning results are shown in Fig. \ref{fig10}, with the results of \emph{DAN(vgg-soft)} shown as dashed lines. We observe that the accuracy first increases and then decreases as $\alpha$ varies and demonstrates a desirable bell-shaped curve. This justifies our assumption that the parameter $\alpha$ in Eqn. (\ref{cluster}) makes a tradeoff between intra-noise and inter-noise of generated pseudo-labels. If $\alpha$ is set to be lower, more faces of different identities will be clustered together which contains severe intra-class noise; otherwise, faces of one identity will split into more scattered points and be discarded, or they will split into smaller clusters leading to severe inter-class noise. 

\begin{figure*}[htbp]
\centering
\includegraphics[width=16cm]{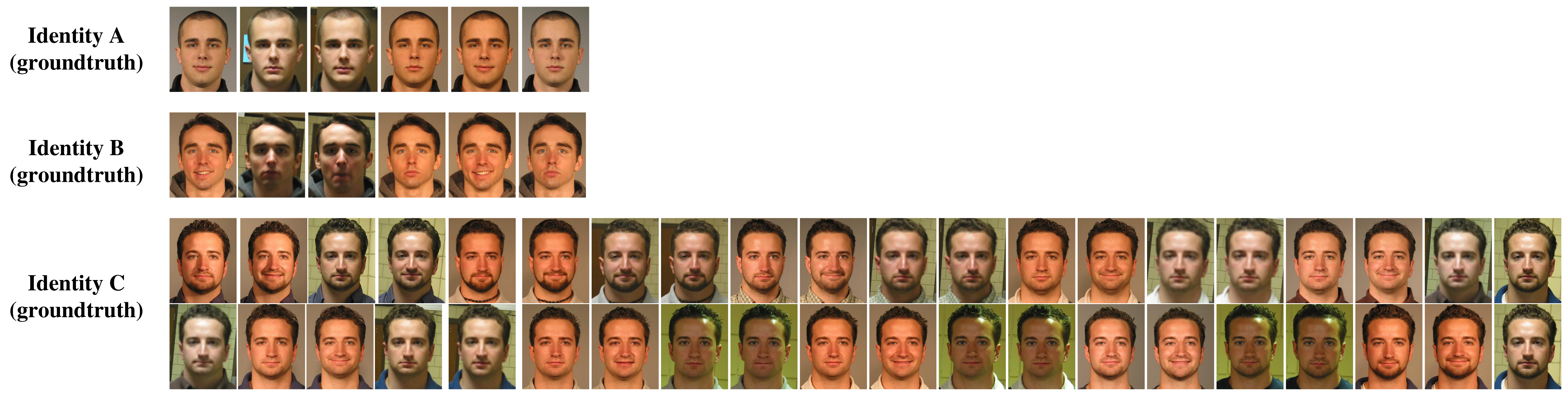}
\caption{ Three examples of ``pure'' clusters generated by clustering method on our target training set of GBU. In top two rows, each row shows the images of one identity; the bottom two rows are images belong to the third identity. For each identity, all images in training set are grouped into one cluster together perfectly.}
\label{fig6}
\end{figure*}

\begin{figure*}[htbp]
\centering
\includegraphics[width=16cm]{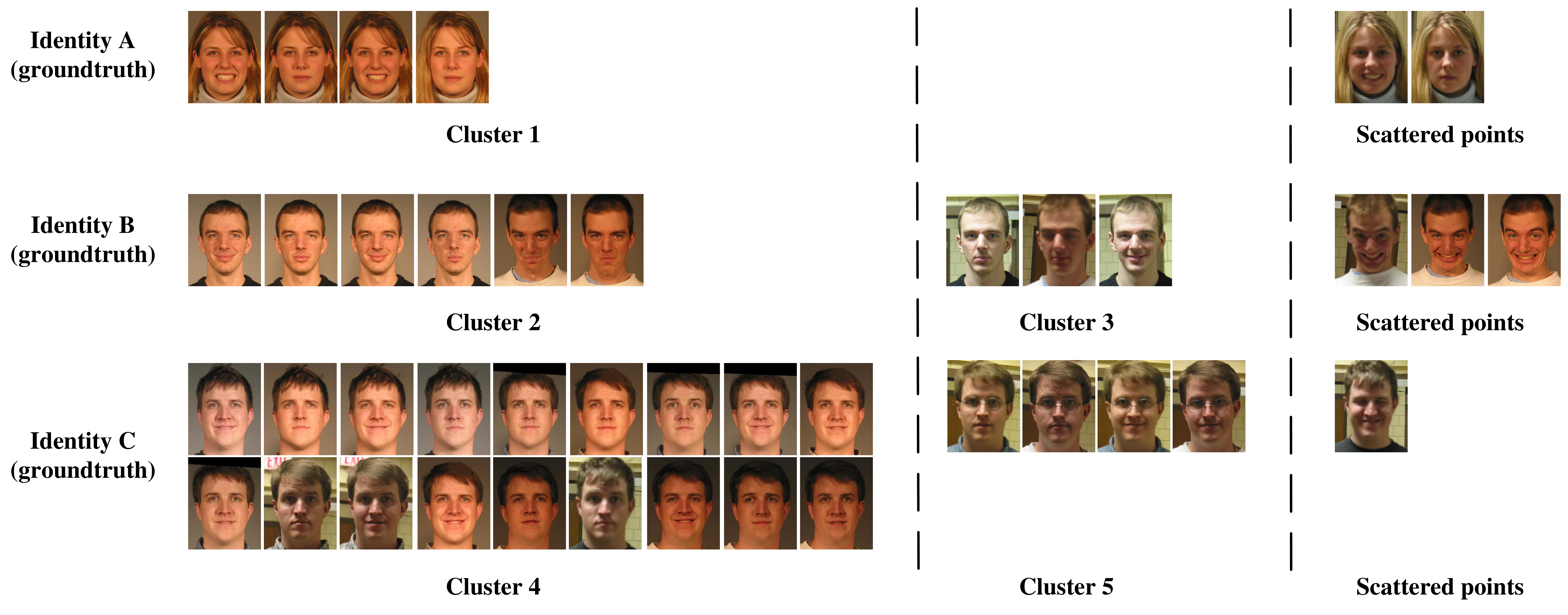}
\caption{ Three examples of ``split'' clusters generated by clustering method on our target training set of GBU. In top two rows, each row shows the images of one identity; the bottom two rows are images belong to the third identity. For the first identity, partial images are clustered together, i.e. $cluster 1$, but remaining images are treated as scattered points and are discarded. For the second and third identity, the images are split into some scattered points and two clusters, which leads to inter-noise.}
\label{fig7}
\end{figure*}

\begin{figure*}[htbp]
\centering
\includegraphics[width=16cm]{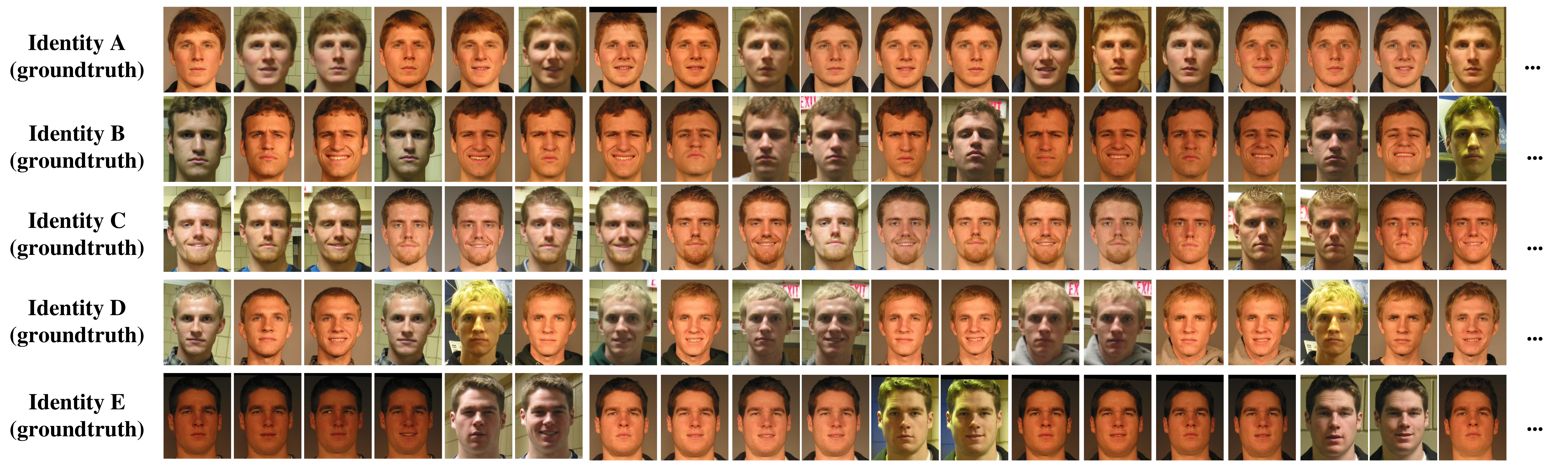}
\caption{ One example of ``impure'' cluster generated by clustering method on our target training set of GBU. Each row shows the images of one identity but all images of these five identities are clustered together incorrectly.}
\label{fig8}
\end{figure*}

\begin{figure}[htbp]
\centering
\includegraphics[width=8cm]{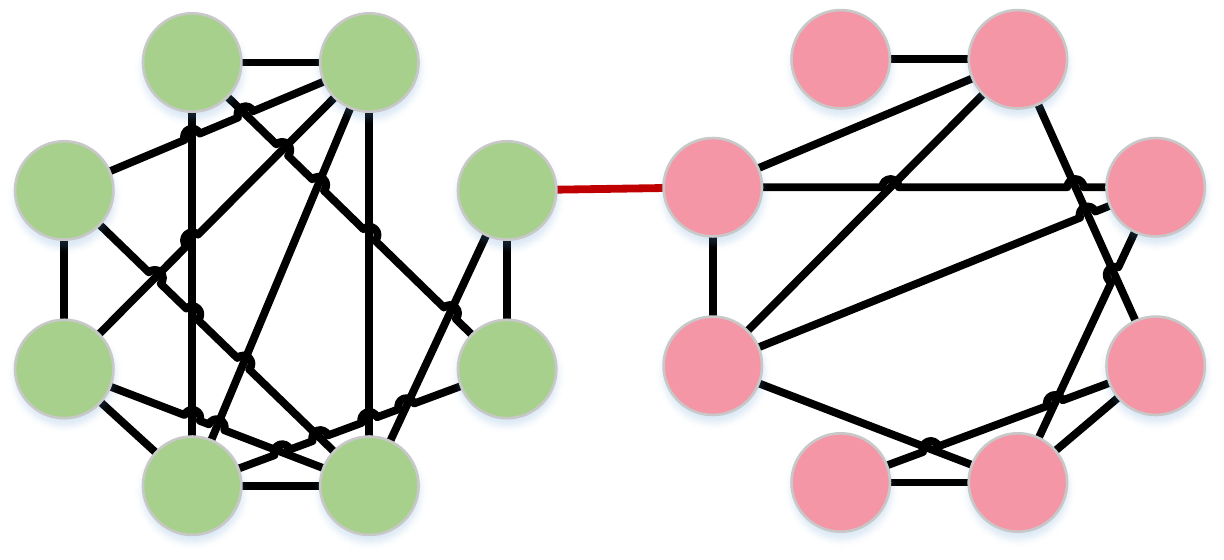}
\caption{ The explanation of the existence of ``impure'' cluster. The green and pink points denote the images of two different identities in one cluster. When there happen to be two or more images of different identities looked like each other, they will be connected in our clustering graph, i.e. the red line, so the images of these people are grouped into one cluster when we generate pseudo-labels through connected component. }
\label{fig9}
\end{figure}

\textbf{Examples of clustering.} As we know, the results of adaptation depend on the quality of pseudo-labels generated by our clustering algorithms. To visually evaluate our clustering method, we show some example clusters on our target training set of GBU in Fig. \ref{fig6}-\ref{fig8}. Fig. \ref{fig6} shows ``pure'' clusters which contain neither intra-noise nor inter-noise, that is to say, all images of one identity are grouped into one cluster together perfectly even if there are variations in expression, lighting, hairstyle, etc. In Fig. \ref{fig7}, examples of ``split'' clusters are presented. Although reliable cluster, e.g. $cluster 2$, is formed with partial images of one identity, remaining images are treated as scattered points or are split into another different clusters, e.g. $cluster 3$, which results in inter-noise. This phenomenon usually occurs due to large variations. Fig. \ref{fig8} shows example of ``impure'' cluster in terms of subject identity. Five different individuals are grouped into one cluster leading to serious intra-noise. When going deep into this type of clusters, we find that it usually happens to the identities whose images' number is quite large. We give the explanation of this phenomenon in Fig. \ref{fig9}. A larger number of images per identity increase the probability of connectivity of different people in our clustering algorithms. Among massive images of two people, there happen to be two or more images of different identities looked like each other and their cosine-similarities are larger than the parameter $\alpha$. Even if two similar images, they will be connected in our clustering graph so the images of these people are grouped into one cluster when pseudo-labels are generated through connected component.

\section{Conclusion}

In this paper, we focus on the issue of domain discrepancy between source training data and target testing data in face recognition scenario. We address it in the viewpoint of unsupervised domain adaptation. First, considering the special problems of non-overlapping classes between two domains in FR, we further propose to introduce clustering algorithms into UDA to obtain pseudo-labels in the deep feature space, and design a simplified spectral clustering algorithm which requires neither overlapping classes between two domains nor the number of target classes. Second, to minimize domain discrepancy and enhance the quality of clustering-based pseudo-labels, we introduce deep UDA methods, namely DDC and DAN. Our CDA method effectively learns the discriminative target feature by aligning the feature domain globally, and, at the meantime, distinguishing the target clusters locally. Comprehensive experiments are carried out in the GBU and IJB-A/B/C databases, significant performance gains are reached which indicates the competency of the proposed approach.

In terms of future work, (1) while the underlying face representation we employ in clustering method works reasonably well for unconstrained face images, it could still be improved in a number of ways (e.g., selecting more reliable source training sets, or improving the transferability of deep model). (2) While we were able to boost the performance of target testing data, the quality of pseudo-labels still needs to be improved. So designing a better clustering method for UDA is a vital problem to be done in FR task. (3) We consider to use the ``easy-to-hard'' scheme which progressively selects reliable pseudo-labeled target samples from the most confident predictions or utilize the training skills of noisy data to alleviate the negative influence of falsely-labeled samples.

{
\bibliographystyle{IEEEtran}
\bibliography{egbib}
}

\end{document}